\documentclass{egpubl}
\usepackage{eg2026}
 
\ConferencePaper        %

\CGFStandardLicense

\usepackage[T1]{fontenc}
\usepackage{dfadobe}  

\usepackage{cite}  %
\BibtexOrBiblatex
\electronicVersion
\PrintedOrElectronic
\ifpdf \usepackage[pdftex]{graphicx} \pdfcompresslevel=9
\else \usepackage[dvips]{graphicx} \fi

\usepackage{egweblnk} 
\usepackage{pdfpages} 

\usepackage[utf8]{inputenc} %
\usepackage[T1]{fontenc}    %
\usepackage{hyperref}       %
\usepackage{url}            %
\usepackage{booktabs}       %
\usepackage{amsfonts}       %
\usepackage{nicefrac}       %
\usepackage{microtype}      %
\usepackage{xcolor}         %
\usepackage{enumitem}
\usepackage{comment}
\usepackage{amsmath}
\usepackage{graphicx}
\usepackage{amssymb}  %
\usepackage{wrapfig}
\usepackage{bm}
\usepackage{newtxtext,newtxmath}
\usepackage{makecell}
\usepackage{multirow}   %

\title{Splat-based Metal Artifact Reduction in Cone-Beam CT\\~via Polychromatic Modeling}

\author[K. Choi, I. Kim, J. Cho, H. Cho, \& M.H. Kim]
{\parbox{\textwidth}{\centering 
Kiseok Choi\orcid{0000-0002-2352-3889} \hspace{5mm} Inchul Kim\orcid{0000-0003-1268-3104} \hspace{5mm} Jaemin Cho\orcid{0000-0003-2800-5105}  \hspace{5mm} Hyeongjun Cho\orcid{0000-0001-9399-4232} \hspace{5mm} Min H. Kim\orcid{0000-0002-5078-4005}}\\
{\parbox{\textwidth}{\centering KAIST}}}

\begin{document}
	
\newcommand{\CHOI}[1]{\TODO{kschoi}{magenta}{#1}}
\newcommand{\JMCHO}[1]{\TODO{jmcho}{orange}{#1}}
\newcommand{\HJCHO}[1]{\TODO{hjcho}{cyan}{#1}}
\newcommand{\KIM}[1]{\TODO{ickim}{violet}{#1}}
\newcommand{\MIN}[1]{\TODO{min}{red}{#1}}

\newcommand{\CHECK}[1]{{\textbf{\color{red}{#1}}}} %

\newcommand{\NEW}[1]{{\color{black}{#1}}} %
\newcommand{\OLD}[1]{{\color{gray}{#1}}}%
\newcommand{\mparagraph}[1]{\vspace{0.5mm}\noindent{\textbf{#1.}\hspace{0.5mm}}}

\def\cites{\textbf{\color{red}{(cites)}}}
\def\XXX{\textbf{\color{red}{XXX}}}

\ifpdf
  \newcommand{\TODO}[3]{\textcolor{#2}{\textbf{[\textsc{#1:} \textit{#3}}]}}
\else
  \newcommand{\TODO}[3]{{\textbf{[\textsc{#1:} \textit{#3}}]}}  %
\fi
\definecolor{cyan}{cmyk}{1,0,0,0}
\definecolor{darkgreen}{rgb}{0,0.5,0}
\definecolor{orange}{rgb}{1,0.5,0}
\definecolor{magenta}{cmyk}{0,1,0,0}
\definecolor{darkyellow}{cmyk}{0,0,0.75,0}
\definecolor{gray}{rgb}{0.8,0.8,0.8}
\newenvironment{tightenumerate}{;
\vspace{-1.5mm}
\begin{enumerate}
  \setlength{\itemsep}{1pt}
  \setlength{\parskip}{0pt}
  \setlength{\parsep}{0pt}}{\end{enumerate}
\vspace{-1.5mm}
}
\newenvironment{tightitemize}{
\vspace{-1.5mm}
\begin{itemize}
  \setlength{\itemsep}{1pt}
  \setlength{\parskip}{2pt}
  \setlength{\parsep}{0pt}}{\end{itemize}

}

\makeatletter
\makeatother

\newcommand{\DELETE}[1]{} %
\newcommand{\IGNORE}[1]{}

\newcounter{datetoday}
\newcounter{diffyears}
\newcounter{diffmonths}
\newcounter{diffdays}

\newcommand{\difftoday}[3]{%
      \setmydatenumber{datetoday}{\the\year}{\the\month}{\the\day}%
      \setmydatenumber{diffdays}{#1}{#2}{#3}%
      \addtocounter{diffdays}{-\thedatetoday}%
      \ifnum\value{diffdays}>0
        \def\diffbefore{}%
        \def\diffafter{left}%
      \else
        \def\diffbefore{}%
        \def\diffafter{ago}%
        \setcounter{diffdays}{-\value{diffdays}}%
      \fi
      \setcounter{diffyears}{\value{diffdays}/365}%
      \setcounter{diffdays}{\value{diffdays}-365*\value{diffyears}}%
      \setcounter{diffmonths}{\value{diffdays}/30}%
      \setcounter{diffdays}{\value{diffdays}-30*\value{diffmonths}}%
      \diffbefore
      \ifnum\value{diffyears}=0
      \else
        \ifnum\value{diffyears}>1
            \thediffyears\space years,
        \else
            \thediffyears\space year,
        \fi
      \fi
      \ifnum\value{diffmonths}=0
      \else
        \ifnum\value{diffmonths}>1
            \thediffmonths\space months
        \else
            \thediffmonths\space month
        \fi
      \fi
      \ifnum\value{diffdays}=0
      \else
        \ifnum\value{diffdays}>1
            \thediffdays\space days
        \else
            \thediffdays\space day
        \fi
      \fi
      \diffafter
}
\def\v{\checkmark}
\def\so{~\\$\bullet$\,~}
\def\o{$\bullet$\,~}
\def\cdm{\,cd/m$^2$}
\newcommand*{\medcap}{\mathbin{\scalebox{1.5}{\ensuremath{\cap}}}}%
\newcommand*{\medcup}{\mathbin{\scalebox{1.5}{\ensuremath{\cup}}}}%

\def\doublehline{\hline\hline}
\def\thickhline{\noalign{\hrule height 1pt}}

	\teaser{
	\vspace{-10mm}
	\centering
	\includegraphics[width=0.95\linewidth]{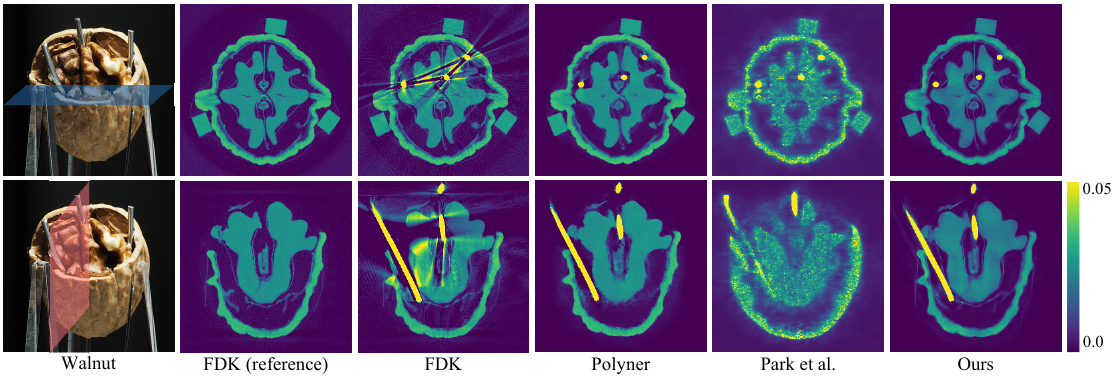}%
	\vspace{-2mm}
	\caption{\label{fig:teaser}
		Qualitative comparison of CBCT reconstruction results for a real walnut object with inserted metal pins. 
		The leftmost column shows the optical photograph of the walnut, where the blue and red planes denote the positions of the horizontal and vertical cuts for visualization.
		Each reconstruction method is visualized with two slices: a horizontal slice (top row) and a vertical slice (bottom row) from the reconstructed volume.
		The FDK (reference) result corresponds to a baseline scan of the walnut \textit{without} metal pins and serves as a ground-truth proxy free from metal artifacts. 
		FDK~\cite{feldkamp1984practical} applied to the metal-inserted scan exhibits severe beam hardening artifacts, including dark streaks and intensity distortions.
		Polyner~\cite{polyner} reduces some artifacts but shows noticeable blurring and oversmoothing of fine structures.
		Park et al.~\cite{yonsei} minimizes artifacts but considerably compromises the overall image structures, suffering from noise.
		Our method delivers the most faithful reconstruction by effectively suppressing artifacts while preserving structural detail in both axial and sagittal views, demonstrating its robustness for cone-beam CT with severe metal-induced beam hardening.}}
	
	\maketitle
\begin{abstract}
Cone-beam computed tomography (CBCT) enables volumetric reconstruction from X-ray projections, but suffers from severe artifacts--especially beam hardening--when imaging materials with high attenuation such as metals. These artifacts arise from the polychromatic nature of X-rays and are not properly addressed by conventional monochromatic reconstruction algorithms. While recent neural representation-based methods offer improved reconstruction quality, they are computationally expensive and often impractical for deployment. We propose a novel physics-inspired, self-calibrating metal artifact reduction method that efficiently reconstructs 3D CBCT volumes while correcting beam hardening artifacts. Our method integrates a polychromatic X-ray projection model, material-dependent attenuation profiles, and system response modeling into a Gaussian Splatting framework. Unlike prior work, we eliminate the need for manual metal masks or strong prior assumptions, and we optimize both reconstruction parameters and X-ray spectral characteristics jointly during training. We further introduce a high-fidelity synthetic CBCT dataset generation pipeline validated on Monte-Carlo x-ray simulation toolbox and release new datasets with severe metal-induced artifacts to support the community. This is the first splat-based method for reducing beam hardening in CBCT. Extensive experiments on both synthetic and real-world datasets demonstrate that our method outperforms state-of-the-art approaches in artifact suppression and reconstruction accuracy.


\end{abstract}

\section{Introduction}

X-rays are high-energy electromagnetic waves capable of penetrating various materials. 
As X-rays propagate through matter, they are attenuated according to the Beer-Lambert law, which models the exponential decay of energy along the path proportional to material density. 
X-ray computed tomography (CT) reconstructs internal volumetric structures by analyzing the attenuation. 
Among several CT configurations—parallel-beam, fan-beam, cone-beam, and helical-beam—cone-beam CT (CBCT) has attracted particular interest due to its ability to capture dense projection views in a single scan via conic emission and 2D planar detection, enabling efficient volumetric acquisition.
This faster acquisition also contributes to reduced X-ray exposure, making CBCT an attractive modality for low-dose tomography applications.

While beam hardening artifacts are a common challenge in most X-ray CT systems due to the polychromatic nature of X-ray sources, where low-energy photons are absorbed more readily than high-energy ones, CBCT is particularly susceptible to such artifacts. 
Its wide cone angle and increased scatter amplify these nonlinear attenuation effects, leading to dark streaks, cupping artifacts, and geometric distortions that significantly degrade image quality (Figure~\ref{fig:teaser}). 
These artifacts not only impair diagnostic interpretation but also hinder downstream computational analysis such as segmentation or 3D modeling.

Traditional beam hardening correction (BHC) methods, including sinogram interpolation using predefined metal masks~\cite{limar, nmar}, are limited by their reliance on heuristic preprocessing.
Data-driven approaches~\cite{adn,dicdnet,acdnet,oscnet} have demonstrated improved artifact removal using supervised learning or dictionary-based priors, but often require metal masks and paired training data, limiting generalization across materials and geometries.
More recently, physically-motivated methods have been proposed to address beam hardening during reconstruction.
Polyner~\cite{polyner} reconstructs energy-dependent voxel attenuations by assuming known X-ray spectra, and Park et al.~\cite{yonsei} leverage a simplified linear attenuation-energy model under fixed energy assumptions.
While these methods show promise, they often depend on prior spectral calibration, operate under restrictive fan-beam geometries, or require computationally expensive neural volumetric representations, making them less suitable for practical CBCT systems.

In this work, we tackle the longstanding challenge of beam hardening artifact reduction in CBCT by introducing a physics-inspired reconstruction framework that models both the polychromatic X-ray spectrum and material-specific attenuation behavior. 
Our method is built upon a differentiable Gaussian Splatting (GS) framework~\cite{r2_gaussian}, and incorporates an energy-aware attenuation model as well as a self-calibrating system response formulation.
This unified framework allows us to jointly reconstruct high-quality attenuation volumes and correct beam hardening artifacts, without requiring any prior knowledge of the X-ray spectrum or manual metal masks.

In summary, we present a physics-inspired reconstruction framework that directly targets the beam hardening problem in cone-beam CT, a long-standing and challenging artifact in low-dose imaging. 
By integrating a polychromatic forward model with material-specific attenuation behavior into a differentiable Gaussian Splatting framework, our method achieves high-fidelity reconstruction while effectively suppressing metal-induced artifacts. 
This is the first splat-based method for reducing beam hardening artifacts in CBCT. 
A key component of our approach is a self-calibrating system response model that eliminates the need for prior spectral information or manual metal masking, significantly improving both robustness and practicality for real-world CBCT systems.

\section{Related Work}

\mparagraph{Beam Hardening Artifact Reduction}
Beam hardening artifacts are a persistent challenge in CT imaging, particularly when scanning high-attenuation materials such as metals. 
The underlying cause is the polychromatic nature of X-ray sources: lower-energy photons are more readily absorbed, leading to nonlinear attenuation that violates the assumptions of monochromatic models and results in dark streaks and geometric distortions.

Early correction methods such as LIMAR~\cite{limar} and NMAR~\cite{nmar} perform sinogram interpolation using predefined metal masks, but suffer from poor generalizability and reliance on accurate mask segmentation.
With the rise of deep learning, data-driven BHC methods have emerged, including ADN~\cite{adn}, InDuDoNet~\cite{indudonet}, DICDNet~\cite{dicdnet}, ACDNet~\cite{acdnet}, and OSCNet~\cite{oscnet}.
These approaches attempt to learn artifact-free mappings but often require paired training data and metal masks, limiting their applicability to varied clinical scenarios.

More recent strategies tackle beam hardening directly during reconstruction by incorporating physical priors.
Polyner~\cite{polyner} reconstructs energy-dependent attenuation volumes assuming a known X-ray spectrum and metal mask, using a NeRF-style representation.
Park et al.~\cite{yonsei} adopt a simplified linear attenuation model and fixed energy assumption, avoiding prior calibration but still relying on heavy NeRF-based volumetric rendering.
While these methods reduce artifacts to some extent, they are computationally expensive and often constrained to fan-beam geometries, which are incompatible with modern CBCT systems.

In contrast, our method targets beam hardening reduction for cone-beam CT by integrating a polychromatic X-ray model and material-specific attenuation profiles into a lightweight Gaussian Splatting framework~\cite{r2_gaussian}. 
Our system includes a self-calibrating component that jointly estimates spectral and attenuation parameters without requiring any prior spectral information or metal masks, offering both generality and practical deployability.

\mparagraph{CT Reconstruction Paradigms}
CT reconstruction algorithms can be broadly categorized into backprojection-based, iterative, and neural volumetric methods.
Backprojection-based methods like FBP~\cite{fbp} and FDK~\cite{feldkamp1984practical} use analytical models rooted in the Fourier slice theorem, but they perform poorly when projection inconsistencies arise, such as those induced by metal artifacts.
Iterative methods~\cite{sart_andersen,sirt} treat reconstruction as an inverse problem and refine the volume estimate through multiple projection-matching steps. 
Edge-preserving regularizations~\cite{asd_pocs,awpcsd,fista} such as total variation improve performance, but they still struggle under severe nonlinearity from beam hardening.
Neural representation approaches like SAX-NeRF~\cite{sax_nerf} apply neural rendering to CT reconstruction, achieving high accuracy but incurring substantial computational cost. 
Variants such as NAF~\cite{naf_zha} and multiresolution hash-based methods~\cite{instantngp_muller} improve runtime at the cost of fidelity.
More recently, Gaussian Splatting (GS)~\cite{gs_original} has been proposed as a neural-free, real-time approach to novel view synthesis.
This method has been adapted to CT in works like R$^2$-Gaussian~\cite{r2_gaussian}, which combines explicit Gaussian representations with the Beer-Lambert law to reconstruct attenuation volumes.
However, R$^2$-Gaussian assumes a monochromatic model and struggles under real-world polychromatic conditions.
Our work builds upon R$^2$-Gaussian by extending it with a physics-inspired polychromatic attenuation model and a self-calibrated system response formulation, enabling artifact-robust cone-beam CT reconstruction under severe beam hardening conditions.

\section{Background}
\subsection{Polychromatic X-ray Projection Model}
\label{subsec:poly_xray_proj}

When an X-ray propagates through an object, its energy is attenuated according to the Beer-Lambert law:
\begin{equation}
	\label{eq:beer_lambert}
	I = I_0 \exp\left( -\int_L \mu(l)\, dl \right),
\end{equation}
where $I_0$ is the emission intensity of the source, $\mu(l)$ is the attenuation coefficient along the line segment~$dl$ of the path~$L$, and~$I$ is the intensity measured at the detector.
This expression can be rearranged into a log-transformed form:
\begin{equation}
	\label{eq:proj_mono}
	P = -\log \frac{I}{I_0} = \int_L \mu(l)\, dl.
\end{equation}
The resulting quantity $P$ is referred to as the projection image or sinogram.
However, since real-world X-ray sources emit a spectrum of energies, and the attenuation coefficient $\mu$ depends on energy, Equation~\eqref{eq:proj_mono} cannot capture the polychromatic nature of X-ray propagation.
To address this, the attenuation process must be expressed as an integral over the energy range~$\mathcal{E}$:
\begin{equation}
	\label{eq:proj_poly}
	P = -\log \int_\mathcal{E} \eta(E) \exp\left( -\int_L \mu(l, E)\, dl \right) dE,
\end{equation}
where~$\eta(E)$ denotes the system response at energy level~$E$, including the X-ray source distribution, filter characteristics, and detector sensitivity. The attenuation coefficient~$\mu(l, E)$ is now energy-dependent and referred to as the linear attenuation coefficient~(LAC).
The measured polychromatic projection is obtained by integrating across the entire spectrum~$\mathcal{E}$.
The LAC is a material-dependent property that varies significantly with energy due to fundamental physical interactions, primarily the photoelectric effect and Compton scattering.
This energy dependence of attenuation serves as a strong prior in estimating material-specific attenuation from polychromatic projections, as described in Equation~\eqref{eq:proj_poly}.

\subsection{R$^2$-Gaussian}

In R$^2$-Gaussian~\cite{r2_gaussian}, the attenuation field in 3D space is modeled as a sum of anisotropic Gaussian primitives:
\begin{equation}
	\label{eq:r2gs_atten}
	\begin{aligned}		
		\mu(\mathbf{x}) = \sum_{i}^{M} \rho_i \exp \left( -\frac{1}{2} (\mathbf{x} - \mathbf{p}_i)^{T} \mathbf{\Sigma}_i^{-1} (\mathbf{x} - \mathbf{p}_i) \right),
	\end{aligned}
\end{equation}
where $\mathbf{x}$ is the spatial query point, $M$ is the number of Gaussians, $\rho_i$ is the density of the $i$-th Gaussian, and~$\mathbf{p}_i$,~$\mathbf{\Sigma}_i$ denote its center and covariance, respectively.
The forward projection of a ray is then computed by integrating the contribution of all Gaussians along the ray path:
\begin{equation}
	\label{eq:r2gs_proj}
	\scalebox{0.85}{$
	\begin{aligned}		
		P \left(\hat{\mathbf{x}}\right) = \int_L \mu(l)\, dl 
		&\approx \sum_{i}^M \sqrt{\frac{2\pi |\tilde{\mathbf{\Sigma}}_i|}{|\hat{\mathbf{\Sigma}}_i|}} \, \rho_i \exp\left( -\frac{1}{2} (\hat{\mathbf{x}} - \hat{\mathbf{p}}_i)^T \hat{\mathbf{\Sigma}}_i^{-1} (\hat{\mathbf{x}} - \hat{\mathbf{p}}_i) \right),
	\end{aligned}
	$}
\end{equation}
where~$\hat{\mathbf{x}}$ is the projected pixel position,~$\tilde{\mathbf{\Sigma}}_i$ is the view-space covariance,~$\hat{\mathbf{\Sigma}}_i$ is the projected covariance, and~$\hat{\mathbf{p}}_i$ is the projected Gaussian center.
The parameters~$\rho_i, \mathbf{p}_i$, and~$\mathbf{\Sigma}_i$ are optimized such that the computed projections from Equation~\eqref{eq:r2gs_proj} match the observed projection images.
After optimization, the final attenuation volume is recovered by evaluating the sum of the fitted Gaussian components.

\section{Method}

\subsection{Overview}
\label{subsec:overview}
\begin{figure*}[t]
	\centering
	\vspace{-3mm}	
	\includegraphics[width=1.0\linewidth]{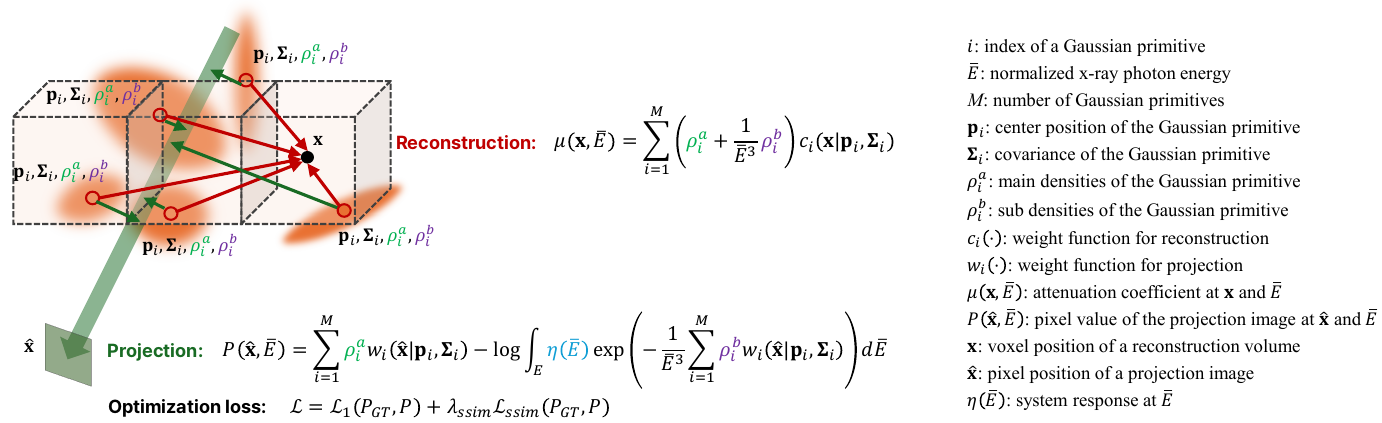}%
	\vspace{-3mm}
	\caption{\label{fig:method_overview}
		Method overview diagram. Our method consists of two types of integrals: projection and reconstruction. In the projection phase, the Gaussians are aggregated using our differentiable polychromatic forward projector, and the resulting projections are compared against ground-truth measurements acquired from the CT system. Importantly, while the per-Gaussian parameters are optimized, the global X-ray system response $\eta(\bar{E})$ is jointly optimized to achieve more effective reduction of metal artifacts. In the reconstruction phase, all optimized Gaussians are integrated to estimate the linear attenuation coefficient at each voxel location, which constitutes the final CT image output. 
	}
	\vspace{-2mm}
\end{figure*}

The overall framework is illustrated in Figure~\ref{fig:method_overview}. Our method consists of two types of integrals: projection and reconstruction. In the projection phase, the Gaussians are aggregated using our differentiable polychromatic forward projector, and the resulting projections are compared against ground-truth log-transformed measurements acquired from the CT system. Importantly, while the per-Gaussian parameters are optimized, the global X-ray system response $\eta(\bar{E})$ is jointly optimized to achieve more effective reduction of metal artifacts. In the reconstruction phase, all optimized Gaussians are integrated to estimate the linear attenuation coefficient at each voxel location, which constitutes the final CT image output.

\subsection{System Response and Polychromatic Attenuation}
\label{subsec:sys_resp}
\mparagraph{System Response}
To accurately model polychromatic X-ray attenuation, it is essential to incorporate the spectral characteristics of the CT system. The system response is defined as the product of the X-ray source spectrum, filter transmission, and detector sensitivity~\cite{xrayphysics}, which are collectively represented by the function~$\eta(E)$. The source spectrum mainly consists of a continuous Bremsstrahlung component with several discrete characteristic peaks~\cite{detector_response}. Since the contribution of these peaks spans only a narrow energy range, they can be neglected in our case. The filter shifts the Bremsstrahlung spectrum toward higher energies~\cite{ahmed2018review}; however, for beam hardening correction, the filter thickness is typically small and its effect can be approximated as negligible. The detector sensitivity decreases gradually with photon energy~\cite{wang2011microct}, which, when combined with the Bremsstrahlung spectrum, results in an approximately linear decay.
At higher energies, material attenuation becomes very weak, so the exponential term $\exp\left( -\int_L \mu(l, E) dl \right)$ approaches unity. Consequently, spectral components in this range cannot be effectively distinguished by gradient-based optimization. Since the spectrum in this region is also close to linearly decaying, we approximate it with a constant plateau, which improves both computational robustness and physical plausibility.
Based on these observations, we approximate the system response using a soft, differentiable piecewise function composed of a linearly decaying component and a constant plateau:
\begin{equation}
	\label{eq:ours_xray_response}
	\scalebox{0.9}{$
		\begin{aligned}
			\eta(\bar{E}) = \ell_{\bar{E}_{th},r}(\bar{E}) \cdot (1 - \sigma(\bar{E} - \bar{E}_{th})) + \ell_{\bar{E}_{th},r}(\bar{E}_{th}) \cdot \sigma(\bar{E} - \bar{E}_{th}),
		\end{aligned}
		$}
\end{equation}
where $\ell_{\bar{E}_{th},r}$ models the linear decay of the response with increasing energy, and $\sigma$ is a sigmoid function that smoothly transitions from the decaying region to a constant value.
For numerical stability and bounded computation, we normalize the energy $E$ (in keV) to $\bar{E}$, which ranges over $[1 - \gamma, 1 + \gamma]$. The parameter~$r$ controls the relative intensity at the threshold energy~$\bar{E}_{th}$ compared to the peak, while~$\gamma$ defines the half-range of the normalized energy domain. Both~$r$ and~$\bar{E}_{th}$ are optimized during reconstruction.
We adopt this simple model for two main reasons: (1) it reduces model complexity and mitigates overfitting, and (2) it decreases the number of estimated parameters, enabling more efficient optimization.

\mparagraph{Polychromatic Attenuation}
To account for the energy-dependent attenuation characteristics of materials, we incorporate the two dominant interaction mechanisms in CBCT energy ranges: the photoelectric effect and Compton scattering, as discussed in Section~\ref{subsec:poly_xray_proj}.
The photoelectric effect dominates at lower energies and exhibits an inverse cubic dependence on energy, while Compton scattering is more prevalent at higher energies~\cite{compton_photoelectric}.
\begin{figure}
	\centering
	\vspace{-3mm}	
	\includegraphics[width=0.7\linewidth]{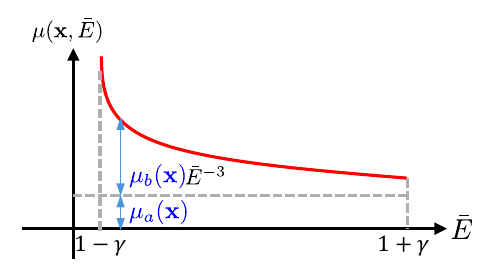}%
	\vspace{-3mm}
	\caption{\label{fig:xray_function_fitting}
		Illustration of our polychromatic attenuation model.
		The material-dependent linear attenuation coefficient $\mu(\mathbf{x}, \bar{E})$ is decomposed into two components: a constant Compton term $\mu_a(\mathbf{x})$ and an energy-dependent photoelectric term $\mu_b(\mathbf{x}) \bar{E}^{-3}$.
		Each Gaussian primitive is assigned these parameters to model spatially varying material attenuation.}
	\vspace{-2mm}
\end{figure}
We model the total attenuation coefficient accordingly:
\begin{equation}
	\label{eq:ours_atten}
	\begin{aligned}		
		\mu(\mathbf{x}, \bar{E}) &= \mu_a(\mathbf{x}) + \frac{1}{\bar{E}^3} \mu_b(\mathbf{x}),
	\end{aligned}
\end{equation}
where $\mu_a(\mathbf{x})$ approximates the Compton scattering component and $\mu_b(\mathbf{x})$ represents the photoelectric effect. The illustration of our model is shown in Figure~\ref{fig:xray_function_fitting}.
To ensure numerical stability during optimization, we approximate the Compton term with a constant value rather than an energy-dependent formulation. This simplification is justified because the Compton component remains nearly constant within the practical X-ray energy range used in medical CT, where the maximum photon energy is below 120 keV~\cite{compton_photoelectric,polyner}. We incorporate this model into the reconstruction framework by assigning each Gaussian primitive a density $\rho_i$ that varies with energy:
\begin{equation}
	\label{eq:ours_atten_gs}
	\begin{aligned}		
		\rho_i(\bar{E}) &= \rho^a_i + \frac{1}{\bar{E}^3} \rho^b_i,
	\end{aligned}
\end{equation}
where~$\rho^a_i$ and $\rho^b_i$ correspond to the terms~$\mu_a$ and~$\mu_b$ in Equation~\eqref{eq:ours_atten} respectively.
Using the Gaussian-based attenuation coefficient~(Equation~\eqref{eq:r2gs_atten}) at~$\bar{E}$ and Equation~\eqref{eq:ours_atten_gs}, we derive Equation~\eqref{eq:ours_atten}:
\begin{equation}
\label{eq:ours_atten_gs_deriv}
	\begin{aligned}		
		\mu(\mathbf{x}, \bar{E}) &= \sum_{i}^{M} {c_i}(\mathbf{x}) \rho_i(\bar{E}) = \sum_{i}^{M} {c_i}(\mathbf{x}) \left( \rho^a_i + \frac{1}{\bar{E}^3} \rho^b_i \right) \\
		&= \underbrace {\sum_{i}^{M} {c_i}(\mathbf{x}) \rho_i^a}_{\triangleq \mu_a(\mathbf{x})} + \frac{1}{\bar{E}^3} \underbrace {\sum_{i}^{M} {c_i}(\mathbf{x}) \rho_i^b}_{\triangleq \mu_b(\mathbf{x})},
	\end{aligned}
\end{equation}
where~${c_i}\left( {\mathbf{x}} \right)= \exp \left( -\frac{1}{2} (\mathbf{x} - \mathbf{p}_i)^{T} \bm{\Sigma}_i^{-1} (\mathbf{x} - \mathbf{p}_i) \right)$ that determines the shape of the~i$^\text{th}$ Gaussian.
Since the attenuation coefficient in Equation~\eqref{eq:ours_atten_gs_deriv} is a multiplication of the geometric Gaussian weight and the density term, it is regularized by the term~${c_i}\left( {\mathbf{x}} \right)$ during optimization.

\subsection{Projection}
We define the polychromatic forward projection model for beam hardening correction.
By substituting the polychromatic attenuation model from Equation~\eqref{eq:ours_atten_gs_deriv} into the forward model~(Equation~\eqref{eq:proj_poly}), we obtain:
\begin{equation}
\label{eq:ours_proj}
\begin{aligned}		
P &= -\log \int_{\bar{\mathcal{E}}} \eta(\bar{E}) \exp\left( -\int_L \mu(l, \bar{E})\, dl \right) d\bar{E} \\
&= -\log \int_{\bar{\mathcal{E}}} \eta(\bar{E}) \exp\left( -\int_L \left( \mu_a(l) + \frac{1}{{\bar{E}}^3} \mu_b(l) \right) dl \right) d\bar{E} \\
&= \int_L \mu_a(l)\, dl - \log \int_{\bar{\mathcal{E}}} \eta(\bar{E}) \exp\left( -\frac{1}{{\bar{E}}^3} \int_L \mu_b(l)\, dl \right) d\bar{E},
\end{aligned}
\end{equation}
where~${\bar{\mathcal{E}}}$ is the integration range in the normalized energy levels (Figure~\ref{fig:xray_function_fitting}).
To compute the projection on the projected coordinate~$\hat{\mathbf{x}}$ using the Gaussian parameters, we plug in Equation~\eqref{eq:r2gs_proj} into Equation~\eqref{eq:ours_proj}:
\begin{equation}
\label{eq:ours_proj_gs}
\begin{aligned}
	P\left( \hat{\mathbf{x}} \right) &\approx \sum_{i}^M w_i\left( \hat{\mathbf{x}} \right) \rho^a_i - \log \int_{\bar{\mathcal{E}}} \eta(\bar{E}) \exp \left( -\frac{1}{\bar{E}^3} \sum_{i}^{M} w_i\left( \hat{\mathbf{x}} \right) \rho^b_i \right) d\bar{E},
\end{aligned}
\end{equation}
where $w_i(\hat{\mathbf{x}}) =
\smash{\sqrt{\scriptstyle 2\pi |\widetilde{\bm{\Sigma}}_i|
		\mathord{\left/ \scriptstyle |\widehat{\bm{\Sigma}}_i| \right.}}}
\exp\!\left(-\frac{1}{2}(\hat{\mathbf{x}} - \hat{\mathbf{p}}_i)^T
\hat{\bm{\Sigma}}_i^{-1}(\hat{\mathbf{x}} - \hat{\mathbf{p}}_i)\right)$
is the weight at the projected coordinate~$\hat{\mathbf{x}}$ of the~i$^\text{th}$ Gaussian primitive.
Note that since we approximate the Compton term as a constant variable~$\mu_a \left(\mathbf{x}\right)$, we can take that term out of the exponential function as shown in the third line of the equation above, which is significantly beneficial in numerical stability during optimization.
To compute~$P\left( \hat{\mathbf{x}} \right)$, we partition the range of normalized energy levels~$\bar{E}$ by~$N$ to discretize Equation~\eqref{eq:ours_proj_gs}:
\begin{equation}
\label{eq:ours_proj_gs_discrete}
\begin{aligned}		
P\left( \hat{\mathbf{x}} \right) &= \sum_{i}^{M} w_i\left( \hat{\mathbf{x}} \right) \rho^a_i\\
&- \log \sum_{k=0}^{N-1} \eta_k \exp \left( - \dfrac{1}{\left( \left( \frac{2k}{N-1} - 1 \right) \gamma + 1 \right)^3} \sum_{i}^{M} w_i\left( \hat{\mathbf{x}} \right) \rho^b_i \right),
\end{aligned}
\end{equation}
where~$\eta_k$ is the normalized system response weight such that~$\sum_{k=0}^{N-1} \eta_k = 1$.

\subsection{Optimization}
We jointly optimize the system response parameters~$r$ and~$\bar{E}_{th}$, along with the Gaussian parameters~$\{ \rho^a_i, \rho^b_i, \mathbf{p}_i, \bm{\Sigma}_i \mid i = 1, \dots, M \}$.
The total loss is a weighted combination of pixel-wise $L_1$ loss and structural similarity (SSIM):
\begin{equation}
	\label{eq:ours_loss}
	\begin{aligned}		
		\mathcal{L}_{total} = \mathcal{L}_1(P_{GT}, P) + \lambda_{ssim} \mathcal{L}_{ssim}(P_{GT}, P),
	\end{aligned}
\end{equation}
where $\lambda_{ssim}$ is fixed to 0.25 in all experiments.
We include the gradient derivation of~$P$ with respect to the Gaussian parameters in the supplemental material to support backpropagation through the R$^2$-Gaussian CUDA pipeline. After optimization, the final voxelized attenuation field is obtained via Equation~\eqref{eq:ours_atten_gs_deriv}.

\subsection{Implementation}
Our method is implemented in PyTorch~\cite{pytorch} with CUDA~\cite{cuda}. We set the normalized energy half-range $\gamma$ to $\frac{(E_{\text{max}} - 10)}{(E_{\text{max}} + 10)}$, assuming that the minimum photon energy in the spectrum is 10~keV and that $E_{\text{max}}$ (the maximum photon energy) is known in advance. 
The number of discrete energy components $N$ used for spectrum integration is fixed to 15.
The parameters for the system response are initialized as $r=0.2$ and $\bar{E}_{th}=1.0$.
The Gaussian parameters in our model are initialized following the same procedure as the base method, R$^2$-Gaussian~\cite{r2_gaussian}. Specifically, we first reconstruct an initial volume using the FDK from the input projection images. From the reconstructed volume, we randomly sample 50,000 positions in the voxel grid where the voxel intensity is greater than 0.005. One Gaussian is then initialized at each of these sampled positions. For each Gaussian, the value of $\rho_{a}$  is initialized by multiplying the voxel intensity by 0.075, and $\rho_{b}$  is initialized with the voxel intensity multiplied by 0.0075. The initialization of the Gaussian covariance follows the same routine used in the original R2-Gaussian algorithm. We adopt R$^2$-Gaussian~\cite{r2_gaussian}'s learning rate scheme, setting $(\rho_a, \rho_b)$ to (0.01, 0.1), and $(r, \bar{E}_{th})$ to 0.001. Learning rates decay exponentially over 20k steps. 

\section{Validation}
\subsection{Synthetic Dataset Generation}
\label{subsec:valid_syn_data_generation}
\begin{figure}[h]
	\centering
	\vspace{-2mm}
	\includegraphics[width=1.0\linewidth]{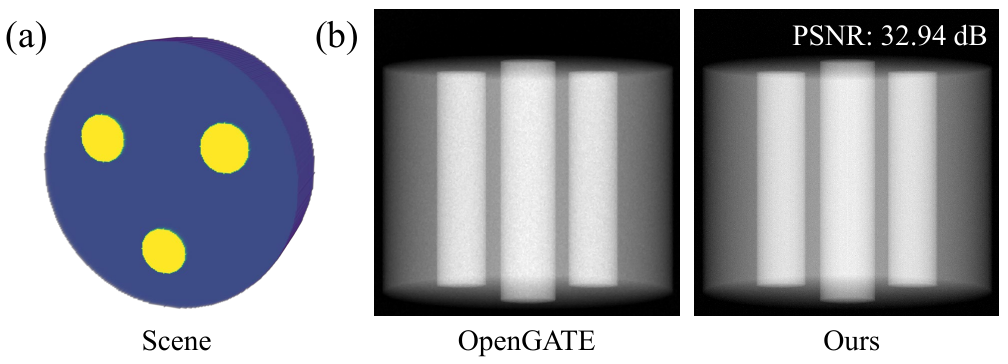}%
	\vspace{-2mm}
	\caption[]{\label{fig:validation_geant4}%
		Validation of our synthetic projection image against one generated by Monte Carlo (MC) simulation using the OpenGATE package. A PSNR of 32.94 dB indicates strong agreement between the two results.}
\end{figure}

To validate our method, we implement polychromatic CBCT synthetic dataset generator based on TIGRE~\cite{tigre_toolbox}, incorporating Poisson and Gaussian noise to mimic realistic acquisition conditions. This simulation pipeline follows established practices in prior work~\cite{lin2019dudonet,adn,zhang2018convolutional}, and we further extend it to the CBCT configuration. For validating our simulator, we employ the OpenGATE package~\cite{opengate} that is based on Geant4 x-ray simulation library which supports Monte-Carlo (MC) projection. We first generate the X-ray spectrum using XrayPhysics library~\cite{kim2023differentiable}. The spectrum ranges between 10--90keV with the take-off angle of 11 degree. The filter consists of aluminum with 0.5mm thickness. We set the material of the detector using GOS (O2SGd2) scintillator with 0.1mm thickness. We set the X-ray source's shape as isotropic to simulate a conic beam, and the half angle of the cone is set to 13 degree. The total number of particles is set to 1 billion for the MC integration. We use a synthetic scene as Figure~\ref{fig:validation_geant4} (a). In the scene, three aluminum rods are inserted into the large cylinder, and the remaining part of the cylinder is filled with water, and the rest of the space consists of air. The volume's physical size is 60 mm $\times$ 60 mm $\times$ 60 mm and the voxelized space has 256$\times$256$\times$256 voxels. We run both our projector and the MC-based one to generate raw detector signals and convert them into projection images by taking negative logarithm. Then, we compute the PSNR between them and get 32.94 dB (Figure~\ref{fig:validation_geant4} (b)), which shows strong agreement of our projection with the MC-based result.

\subsection{System Response Model Selection}
We adopt the simple yet physics-inspired approximation model for the system response described in Section~\ref{subsec:sys_resp}, which has only two parameters. To validate the superiority of this choice, we evaluate reconstruction performance using more flexible alternatives, including a Gaussian Mixture Model (GMM) and a model with free parameters. As shown in Table~\ref{tb:ablation_system_response}, our model outperforms these alternatives.

\begin{table}[h]
	\centering
	\caption{\label{tb:ablation_system_response}
		Validation of our system response model against alternative flexible models (scene: Lung).}	
	\resizebox{0.7\linewidth}{!}{%
		\begin{tabular}{lcc}
			\toprule
			System response & PSNR3D & SSIM3D \\
			\midrule
			GMM(w/ 5 Gaussians)              & 26.51 & 0.989 \\
			Free parameters  & 27.74 & 0.991 \\
			Our model  & \textbf{29.19} & \textbf{0.993} \\
			\bottomrule
		\end{tabular}
	}%
\end{table}

\subsection{Generalizability to Multiple CT Systems}
We validate the robustness of our method using synthetically generated projections that emulate different detector types and ranges of X-ray energies, as shown in Table~\ref{tb:ablation_ct_systems}. Our method shows consistent performance across various hardware configurations.

\begin{table}[h]
	\centering
	\caption{\label{tb:ablation_ct_systems}Validation of our method’s generalizability under various acquisition conditions (scene: Lung).}
	\resizebox{1.0\linewidth}{!}{%
		\begin{tabular}{cccccc}
			\toprule
			Case & Voltage (kV) & Filter (mm) & Detector & PSNR3D & SSIM3D \\
			\midrule
			1 & 90  & Al 0.5 & GOS & 29.12 & 0.993 \\
			2 & 90  & Al 1.0 & GOS & 30.12 & 0.994 \\
			3 & 90  & Al 0.5 & CsI & 29.38 & 0.993 \\
			4 & 120 & Al 0.5 & GOS & 28.91 & 0.990 \\
			5 & 60  & Cu 0.5 & GOS & 31.17 & 0.995 \\
			\bottomrule
		\end{tabular}
	}%
	\vspace{-3mm}
\end{table}

\subsection{Real Attenuation Estimation}
\begin{figure}[h]
	\centering
	\includegraphics[width=1.0\linewidth]{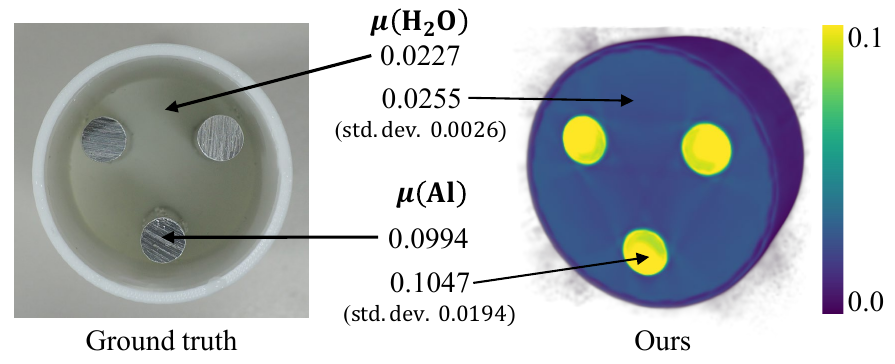}%
	\vspace{-2mm}
	\caption[]{\label{fig:results_ablation_nist}%
		Validation of reconstructed attenuation coefficients against the NIST database~\cite{nist}. All values are reported in $\text{mm}^{-1}$}
\end{figure}

In the real scene \textit{Metal Rods} (the third row in Figure~\ref{fig:results_real_comparison}), water and aluminum rods are used. After running our reconstruction, we segment the reconstructed volume into regions corresponding to water and aluminum via thresholding. We then compare the reconstructed attenuation values for each segment with the ground-truth linear attenuation coefficients of water and aluminum from the NIST database~\cite{nist} (at the center energy 50keV). As shown in the Figure~\ref{fig:results_ablation_nist}, the estimated values are well aligned with the real-world attenuation coefficients in the homogeneous regions.

\section{Results}
\subsection{Dataset}
\label{subsec:dataset}
\begin{figure}[h]
	\centering
	\includegraphics[width=1.0\linewidth]{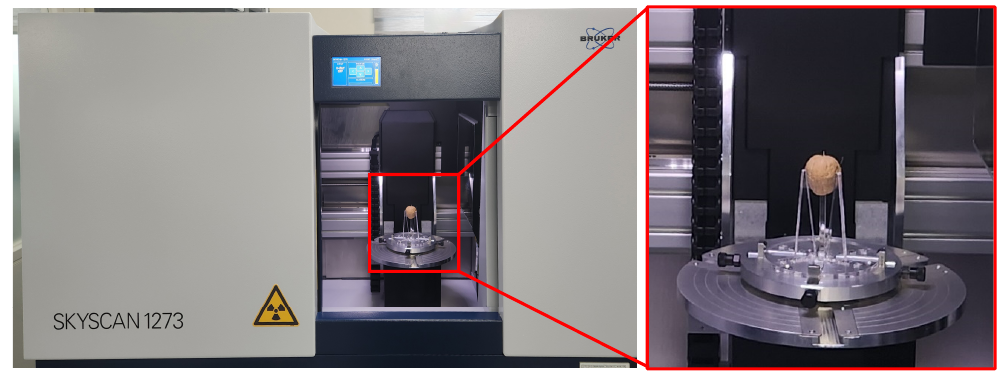}%
	\vspace{-2mm}
	\caption[]{\label{fig:system}%
		Bruker SKYSCAN 1273 CBCT system.}
\end{figure}
We validate our method both on synthetic datasets and real datasets.
For the synthetic evaluation, we utilize the LIDC dataset~\cite{lidc}, the X-plant dataset~\cite{xplant}, and the ZCB100 dataset~\cite{zcb100}. 
We evaluate our method on three synthetic CBCT phantoms (\textit{Lung}, \textit{Teeth}, and \textit{Broccoli}) and five real-world scans: a walnut with metal pins (\textit{Walnut}), a metal-embedded cylinder (\textit{Metal Rods}), a chicken wrapped in metal wire (\textit{Chicken}), a bell pepper with metal rivets (\textit{Bell Pepper}), and a broccoli with metal rivets (\textit{Broccoli}). 
Synthetic projections are generated using our validated simulator introduced in Section~\ref{subsec:valid_syn_data_generation}.
Real projections are acquired using the Bruker SKYSCAN 1273 CBCT system (Figure~\ref{fig:system}) at a tube voltage of 90~kVp with a 0.5~mm aluminum filter. Each scan consists of 720 projections at a resolution of $512 \times 512$.
We scan each specimen both with and without inserted metal materials to observe metal-induced beam hardening effects.
To validate the reconstruction accuracy of our method on real objects, we fabricated a 3D-printed cylindrical phantom containing three aluminum alloy rods to induce severe artifacts (see the lower left of  Figure~\ref{fig:results_real_comparison}). 
Figure~\ref{fig:results_real_comparison} (third column) shows the volume reconstruction exhibiting prominent beam hardening patterns. 
All projection images are resized to the resolution of $512 \times 512$, and each scene has 720 projections.

\subsection{Synthetic Experiment}
\begin{figure*}[h]
	\centering
	\includegraphics[width=1.0\linewidth]{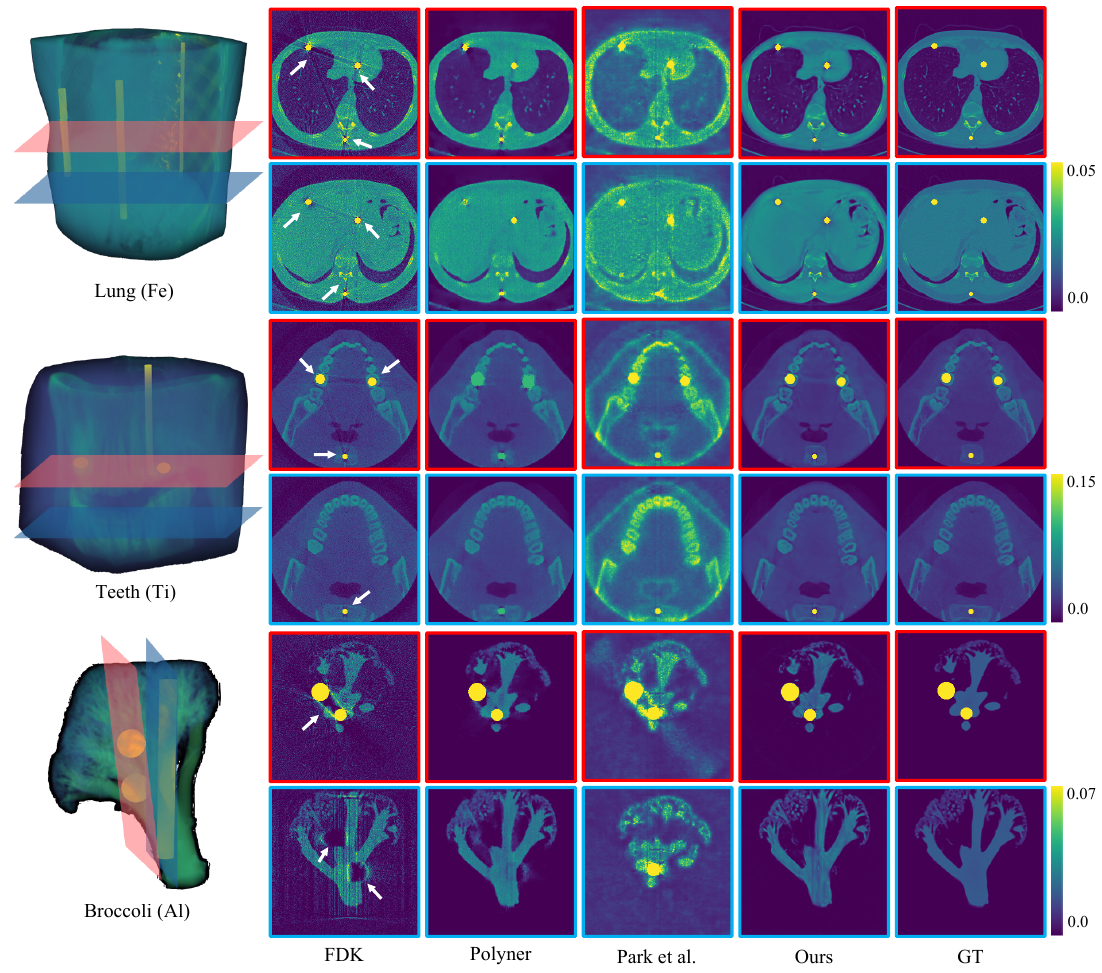}%
	\vspace{-3mm}
	\caption{\label{fig:results_synthetic_comparison_body1}
		Qualitative comparison on synthetic CBCT datasets.
		Reconstruction results on Lung, Teeth, and Broccoli scenes with metallic implants of different materials. Our method achieves superior artifact suppression and structural fidelity compared to prior methods.}
	\vspace{+2mm}
\end{figure*}

\begin{figure*}[h]
	\centering
	\includegraphics[width=1.0\linewidth]{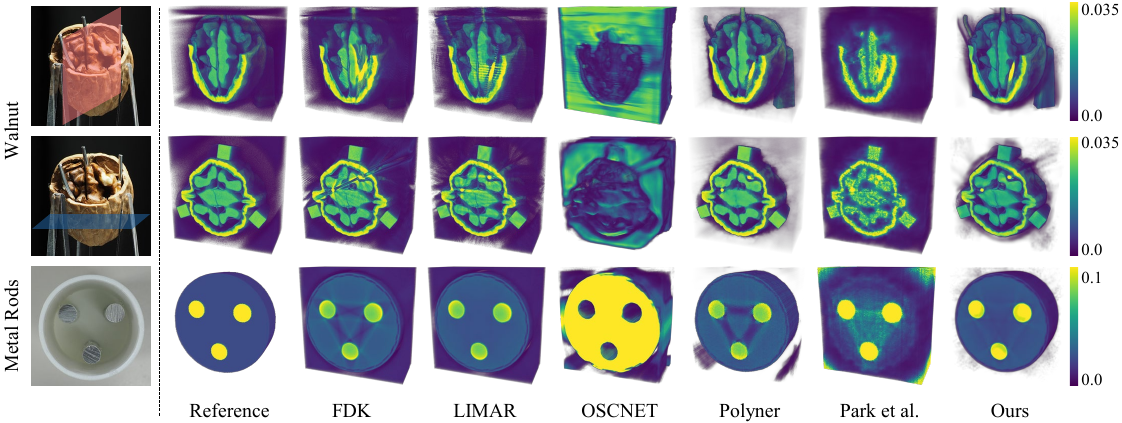}%
	\vspace{-3mm}
	\caption{\label{fig:results_real_comparison}
		Qualitative comparison on real CBCT datasets.
		Reconstruction results for walnut and \textit{Metal Rods} phantoms. 
		Reference volumes are obtained from either separate scans or synthetic generation. 
		Our method achieves clearer structures and stronger artifact suppression compared to prior approaches.	}
	\vspace{-4mm}
\end{figure*}

We evaluate the performance of our method on three synthetic CBCT scenes—\textit{Lung}, \textit{Teeth}, and \textit{Broccoli}—each containing embedded metallic objects made of iron (Fe), titanium (Ti), and aluminum (Al), respectively. 
The volumes are constructed with anatomical and organic structures, and cylindrical or spherical metal implants are synthetically inserted to induce beam hardening artifacts.
CBCT projections are simulated under realistic polychromatic conditions using XrayPhysics~\cite{xrayphysics}.
Figure~\ref{fig:results_synthetic_comparison_body1} compares the reconstruction quality of various methods on these datasets.
FDK~\cite{feldkamp1984practical} exhibits severe streaking artifacts, even in slices not directly intersected by metal due to 3D projection effects inherent in cone-beam geometry.
While methods such as LIMAR~\cite{limar}, NMAR~\cite{nmar}, and Polyner~\cite{polyner} partially mitigate artifacts, they still suffer from residual distortions or blurring.
Park et al.~\cite{yonsei}, originally developed for fan-beam CT, struggle to generalize to cone-beam settings, resulting in poor reconstruction quality.
In contrast, our method produces high-fidelity reconstructions with accurate volume recovery and significantly suppressed metal-induced artifacts across all scenes.
Table~\ref{tb:synthetic_experiment} quantitatively reports 3D PSNR and SSIM metrics.
Our method consistently achieves the highest scores in all scenes, surpassing existing classical, learning-based, and physics-guided baselines.
These results highlight the effectiveness of jointly modeling polychromatic attenuation and system response within a physics-inspired reconstruction framework.
\begin{table}[ht]
	\centering
	\caption{\label{tb:synthetic_experiment}
	Quantitative results on synthetic CBCT datasets.
3D PSNR and SSIM scores for the Lung, Teeth, and Broccoli scenes corresponding to Figure~\ref{fig:results_synthetic_comparison_body1}. 
Our method achieves the highest accuracy across all datasets, demonstrating strong artifact suppression and structural fidelity.}
	\resizebox{1.0\linewidth}{!}{%
	\begin{tabular}{l|cc|cc|cc}
		\hline
		& \multicolumn{2}{c|}{Lung} & \multicolumn{2}{c|}{Teeth} & \multicolumn{2}{c}{Broccoli} \\
		\cline{2-7}
		Method & PSNR3D & SSIM3D & PSNR3D & SSIM3D & PSNR3D & SSIM3D \\
		\hline
		FDK~\cite{feldkamp1984practical} & 19.04 & 0.924 & 32.84 & 0.917 & 11.66 & 0.892 \\
		LIMAR~\cite{limar}               & 19.83 & 0.935 & 32.97 & 0.921 & 13.60 & 0.922 \\
		NMAR~\cite{nmar}                 & 20.49 & 0.942 & 33.05 & 0.923 & 15.81 & 0.951 \\
		ACDNet~\cite{acdnet}             & 13.88 & 0.644 & 26.28 & 0.602 & 14.37 & 0.943 \\
		DICDNet~\cite{dicdnet}           & 13.48 & 0.605 & 25.33 & 0.423 & 13.93 & 0.933 \\
		OSCNet~\cite{oscnet}             & 13.28 & 0.592 & 25.34 & 0.422 & 13.83 & 0.931 \\
		Polyner~\cite{polyner}           & 20.43 & 0.955 & 30.18 & 0.959 & 18.88 & 0.987 \\
		Park et al.~\cite{yonsei}        & 13.59 & 0.699 & 23.15 & 0.524 & 5.29  & 0.526 \\
		Ours & \textbf{29.19} & \textbf{0.993} & \textbf{35.97} & \textbf{0.992} & \textbf{23.22} & \textbf{0.995} \\
		\hline
	\end{tabular}
	\vspace{-5mm}
}
\end{table}

\begin{figure*}[h]
	\vspace{-2mm}
	\centering
	\includegraphics[width=1.0\linewidth]{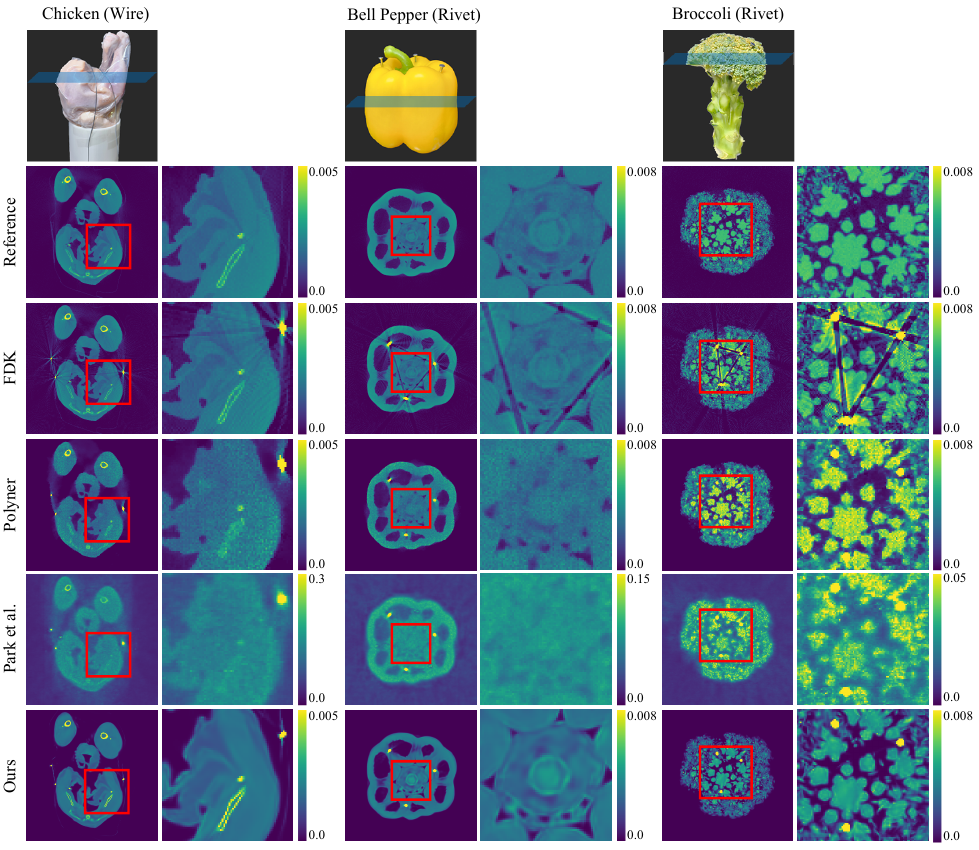}%
	\vspace{-2mm}
	\caption{\label{fig:results_real_comparison_3objects}
		Qualitative comparison on real CBCT datasets.
		The slicing locations are displayed on each picture of the real objects in the first row.
		Odd columns show representative slices reconstructed by each method, with red boxes highlighting regions of interest. Even columns display the corresponding zoom-in patches. 
		Additional slices are provided in the supplemental document. 
		Our method produces clearer reconstructions with fewer artifacts compared to prior approaches.
	}
	\vspace{-4mm}
\end{figure*}

\subsection{Real Experiment}
To evaluate the practical applicability of our method, we conduct experiments on real CBCT measurements acquired with the aforementioned X-ray scanner. We test five physical scenes as described in Section~\ref{subsec:dataset}. All scenes except \textit{Metal Rods} are also scanned separately without metal for reference, while the artifact-free version of the cylindrical phantom is synthesized in simulation for ground-truth comparison. Note that the outer shell of the cylindrical phantom is made of polylactide, whose attenuation is very similar to that of water.
Figure~\ref{fig:results_real_comparison} presents qualitative comparisons across multiple reconstruction methods. FDK and conventional algorithms (e.g., LIMAR~\cite{limar}, NMAR~\cite{nmar}) exhibit severe streaking and low-frequency distortions that obscure important structures in both axial and sagittal views. The data-driven method~\cite{oscnet} fails to generalize to volume reconstruction due to out-of-distribution issues. Polyner~\cite{polyner} achieves reasonable reconstructions but leaves residual artifacts, while Park et al.~\cite{yonsei} struggles to adapt to the CBCT setting. In contrast, our method consistently produces volumetrically accurate reconstructions with sharp structural boundaries and substantially fewer artifacts across objects. These results highlight the robustness and real-world transferability of our framework, even under hardware and acquisition conditions not seen during development.
Figure~\ref{fig:results_real_comparison_3objects} further shows results for three organic objects across joint reconstruction and metal artifact reduction methods. Odd columns display representative slices reconstructed by each method, with red boxes marking regions of interest; even columns show the corresponding zoom-in patches. Additional slices are provided in the supplemental material. Our method produces clearer reconstructions with fewer artifacts compared to prior approaches.

\subsection{Computation Time}
We compare the computation time of our method with state-of-the-art joint reconstruction and metal artifact correction approaches in Table~\ref{tb:ablation_computation_time}. All measurements are performed on an Intel Xeon 4214R CPU and an NVIDIA RTX A6000 GPU. Our method achieves an order-of-magnitude speedup across all scenes.
\begin{table}[h]
	\centering
	\caption{\label{tb:ablation_computation_time}
		Comparisons of computation time. Our method is faster than the other joint reconstruction and metal artifact reduction methods.
	}
	\renewcommand{\arraystretch}{1.4} %
	\resizebox{0.9\linewidth}{!}{%
		\begin{tabular}{c|c|r r r}
			\hline
			& Scene & \multicolumn{1}{c}{Polyner~\cite{polyner}} & \multicolumn{1}{c}{Park et al.~\cite{yonsei}} & \multicolumn{1}{c}{Ours} \\
			\hline
			\multirow{3}{*}{\rotatebox[origin=c]{90}{\parbox{1.5cm}{\centering Synthetic}}} 
			& Broccoli & 1h 35m 58s & 1h 33m 41s & \textbf{16m 08s} \\
			& Lung     & 1h 18m 19s & 1h 15m 35s & \textbf{22m 16s} \\
			& Teeth    & 1h 19m 50s & 1h 18m 34s & \textbf{23m 38s} \\
			\hline
			\multirow{5}{*}{\rotatebox[origin=c]{90}{\parbox{1.2cm}{\centering Real}}} 
			& Walnut          & 1h 11m 58s & 1h 10m 20s & \textbf{17m 51s} \\
			& Metal Rods      & 1h 27m 14s & 1h 37m 56s & \textbf{30m 41s} \\
			& Broccoli        & 2h 06m 55s & 1h 59m 16s & \textbf{16m 16s} \\		
			& Bell Pepper     & 2h 01m 00s & 2h 10m 11s & \textbf{15m 33s} \\					
			& Chicken     	  & 2h 02m 55s & 2h 09m 43s & \textbf{16m 35s} \\
			\hline
		\end{tabular}
	}%
	\vspace{-5mm}
\end{table}

\section{Ablation Study}
\subsection{Component Contribution}
To evaluate the effectiveness of each core component in our framework, we perform an ablation study on the \textit{Lung} scene by systematically adding the proposed modules to the baseline R$^2$-Gaussian algorithm. 
Specifically, we investigate the individual and combined impacts of the system response model and the polychromatic attenuation model.
As shown in Table~\ref{tb:ablation_component}, incorporating the system response model alone yields a modest improvement in both PSNR and SSIM, suggesting that response-aware weighting improves projection fidelity. 
In contrast, the addition of the polychromatic attenuation model leads to a substantial gain in reconstruction quality, highlighting the importance of accurate energy-dependent material modeling in suppressing beam hardening artifacts.
When both modules are integrated, we observe the highest reconstruction performance, achieving a PSNR of 29.19 and an SSIM of 0.993, thereby confirming the complementary nature of the two components.
\begin{table}[h]
	\centering	
	\vspace{-3mm}
	\caption{\label{tb:ablation_component}
		Component contribution analysis. We examine the individual contributions of the system response model and polychromatic attenuation model. 
		Both components contribute positively to reconstruction accuracy, with their combination yielding the best overall performance (scene: Lung).}
	\resizebox{0.8\linewidth}{!}{%
		\begin{tabular}{l|cc}\hline
			Module & PSNR3D & SSIM3D \\ 
			\hline
			Baseline & 20.68 & 0.954 \\
			Baseline with response model & 21.21 & 0.963 \\
			Baseline with attenuation model & 27.73 & 0.991 \\		
			Baseline with both & \textbf{29.19} & \textbf{0.993} \\					\hline
		\end{tabular}
		}%
	\vspace{-3mm}
\end{table}

\subsection{Number of Spectrum Components}
We conduct an experiment to determine the optimal number of spectrum components ($N$). As shown in Table~\ref{tb:ablation_spectrum_count}, the best reconstruction quality is achieved when $N = 15$. 
When $N$ exceeds 15, the performance starts getting saturated and slightly degraded.%
\begin{table}[h]
	\centering	
	\vspace{-1mm}
	\caption{\label{tb:ablation_spectrum_count}
		Performance impact of the number of spectral components. The best performance is achieved at $N=15$ (scene: Lung).
	}
	\resizebox{0.6\linewidth}{!}{%
		\begin{tabular}{c|cc}
			\hline
			\makecell{\# of Spectrum\\Components~(N)} & PSNR3D & SSIM3D \\ 
			\hline
			7 & 27.86 & 0.988 \\
			15 & \textbf{29.19} & \textbf{0.993} \\
			31 & 28.70 & \textbf{0.993} \\		
			63 & 28.25 & 0.992 \\					
			\hline
		\end{tabular}
	}%
	\vspace{-5mm}
\end{table}

\subsection{Performance Comparison to Baseline}
We compare our method against the baseline approach~\cite{r2_gaussian} on scenes containing metal artifacts.
By incorporating a polychromatic attenuation model and system response, our method consistently achieves superior performance across different scenes, as shown in Table~\ref{tab:ablation_baseline_comparison}.
\begin{table}[h]
	\centering
	\vspace{-2mm}
	\caption{Comparison of the baseline~\cite{r2_gaussian} and our method on PSNR3D and SSIM3D across different scenes.}
	\resizebox{0.8\linewidth}{!}{%
	\begin{tabular}{lcccc}
		\toprule
		Scene & \multicolumn{2}{c}{Baseline~\cite{r2_gaussian}} & \multicolumn{2}{c}{Ours} \\
		(synthetic)& PSNR3D & SSIM3D & PSNR3D & SSIM3D \\
		\midrule
		Broccoli & 17.67 & 0.984 & \textbf{23.22} & \textbf{0.995} \\
		Lung     & 20.68 & 0.954 & \textbf{29.19} & \textbf{0.993} \\
		Teeth    & 33.53 & 0.964 & \textbf{35.97} & \textbf{0.992} \\
		\bottomrule
	\end{tabular}
	}%
	\label{tab:ablation_baseline_comparison}
\end{table}
We also compare the computation time between our method and the baseline~\cite{r2_gaussian} as Table~\ref{tab:time_comparison_baseline}.
Due to the additional computation of the polychromatic model in our forward projection compared to the baseline method, each optimization iteration takes slightly longer. To mitigate this overhead, we implement several efficiency-oriented strategies: we analytically simplify the forward projection and its corresponding backward gradient computations, and eliminate redundant operations to reduce computation costs. For faster convergence, we conduct extensive experiments to identify optimal learning rates for $\rho_{a}$ and $\rho_{b}$. As a result, despite the per-iteration overhead, our model converges within 20K iterations, whereas the baseline method typically requires around 30K. This demonstrates that, with proper hyperparameter tuning, our model achieves comparable or even faster convergence.
\begin{table}[h]
	\centering
	\vspace{-3mm}
	\caption{Convergence time comparison between the baseline and our method across different scenes.}
	\resizebox{0.8\linewidth}{!}{%
	\begin{tabular}{llcc}
		\toprule
		Type & Scene & Baseline~\cite{r2_gaussian} & Ours \\
		\midrule
		\multirow{3}{*}{Synthetic}
		& Broccoli         & 18m 44s & 16m 08s \\
		& Lung             & 20m 17s & 22m 16s \\
		& Teeth            & 21m 35s & 23m 38s \\
		\midrule
		\multirow{5}{*}{Real}
		& Walnut           & 20m 52s & 17m 51s \\
		& Metal Rods       & 27m 07s & 30m 41s \\
		& Broccoli & 17m 52s & 16m 16s \\
		& Bell Pepper & 19m 53s & 15m 33s \\
		& Chicken   & 21m 59s & 16m 35s \\
		\bottomrule
	\end{tabular}
	}%
	\vspace{-3mm}
	\label{tab:time_comparison_baseline}
\end{table}

\subsection{Sensitivity on $\gamma$}
We normalize the energy spectrum so that its center becomes 1 and the minimum and maximum values are set to $1-\gamma$ and $1+\gamma$, respectively. If $\gamma$ is not set correctly and does not match the actual spectrum range, reconstruction quality may deteriorate. We evaluate the sensitivity by running our method using projections (acquired under a 10--90 keV spectrum) with various $\gamma$ values corresponding to maximum energies of 60, 90, 120, and 150 keV. As shown in Table~\ref{tab:gamma_sensitivity}, the model shows the best performance when $\gamma$ is set near the true spectrum range (i.e., 90 keV), which validates the significance of the choice of~$\gamma$.
\begin{table}[h!]
	\centering
	\vspace{-3mm}
	\caption{Effect of varying $\gamma$ and corresponding maximum energy $E_{\text{max}}$ on reconstruction quality (scene: Broccoli).}
	\label{tab:gamma_energy_sweep}	
	\resizebox{0.8\linewidth}{!}{%
	\begin{tabular}{cccc}
		\toprule
		$\gamma$ & Corresponding $E_{\text{max}}$ (keV) & PSNR3D & SSIM3D \\
		\midrule
		0.714 & 60  & 21.85 & 0.9943 \\
		0.800 & 90  & \textbf{23.22} & \textbf{0.9953} \\
		0.846 & 120 & 22.83 & 0.9946 \\
		0.875 & 150 & 22.38 & 0.9939 \\
		\bottomrule
	\end{tabular}
	}%
	\vspace{-3mm}
	\label{tab:gamma_sensitivity}
\end{table}

\section{Discussion}
\label{sec:discussion}
Due to limitations in accessing clinical CT datasets, we were unable to conduct comprehensive evaluations across diverse real medical use cases. Instead, we have validated our method on synthetic human-body datasets containing metal implants (e.g., in soft tissue and dental regions). Additionally, we have tested on a real-world dataset where metallic wires were inserted into a chicken to induce metal artifacts in biological tissues. These experiments show that our method generalizes well to various anatomical objects and metals. We are actively seeking access to clinical data to further investigate the generalizability of our method under real-world scenarios for future work.

While our method is built upon a physics-inspired polychromatic projection model, its assumptions may break down under extreme conditions such as photon starvation or partial volume effects. Also, it relies on approximations of system response and material attenuation models. As a result, the reconstructed linear attenuation coefficients may deviate from the true physical values for specific materials.

To enable fair comparison, we re-implemented Polyner~\cite{polyner} and Park et al.~\cite{yonsei}, as no CBCT-compatible public code was available. After careful tuning, our Polyner reproduction matched or exceeded the original performance in CBCT settings. In contrast, Park et al.’s fan-beam-based method showed limited generalization to CBCT, highlighting the challenge of adapting fan-beam approaches to cone-beam geometries under severe artifacts. Details of the comparison implementation are provided in the supplemental document.

\section{Conclusion}
\label{sec:conclusion}
We have presented a CBCT reconstruction method that reduces metal-induced beam hardening artifacts via polychromatic modeling and Gaussian splatting. 
Our framework jointly estimates volume structure, system response, and attenuation parameters—without requiring metal masks, paired supervision, nor spectral priors.
Extensive evaluations on synthetic and real data confirm that our method outperforms state-of-the-art baselines in reconstruction accuracy and artifact suppression. 
We also release new CBCT datasets with realistic artifacts to support future research in artifact-resilient reconstruction.

\section*{Acknowledgements}
\label{sec:acknow}
Min H.~Kim acknowledges the Samsung Research Funding \& Incubation Center (SRFC-IT2402-02), the Korea NRF grant (RS-2024-00357548), the MSIT/IITP of Korea (RS-2022-00155620, RS-2024-00398830, RS-2024-00436680, and 2017-0-00072), and Microsoft Research Asia.

\bibliographystyle{eg-alpha-doi} 
\bibliography{bibliography}       

\clearpage 
\pagestyle{empty} 


\section*{Supplementary material}

# Supplementary Material: Splat-based Metal Artifact Reduction in Cone-Beam CT via Polychromatic Modeling

Kiseok Choi Inchul Kim Jaemin Cho Hyeongjun Cho Min H. Kim

KAIST

## 1. Overview of Our Optimization Pipeline

We illustrate our overall optimization pipeline in Figure 1. Our method consists of two stages, following a structure similar to the baseline approach [ZLC*24]. In the first stage, a conventional reconstruction method, such as FDK [FDK84], is employed to obtain an initial volumetric reconstruction. From this volume, we randomly sample a subset of voxel locations and initialize 3D Gaussian primitives with their centers placed at the sampled positions. These Gaussians are forward-projected onto the projection image domain under a cone-beam geometry using the given system response parameters. The resulting synthetic projections are compared with the measured projection images, and the reconstruction loss is backpropagated to jointly optimize the corresponding 3D Gaussian parameters and the system response parameters. During this iterative optimization, our adaptive density control strategy is applied to prune the negligible Gaussians and to split or clone Gaussians in order to better represent the underlying volume. The output of the first stage is a set of optimized 3D Gaussian parameters. In the second stage, the optimized Gaussians are evaluated at the voxel locations of interest, from which the final reconstructed volume is obtained.

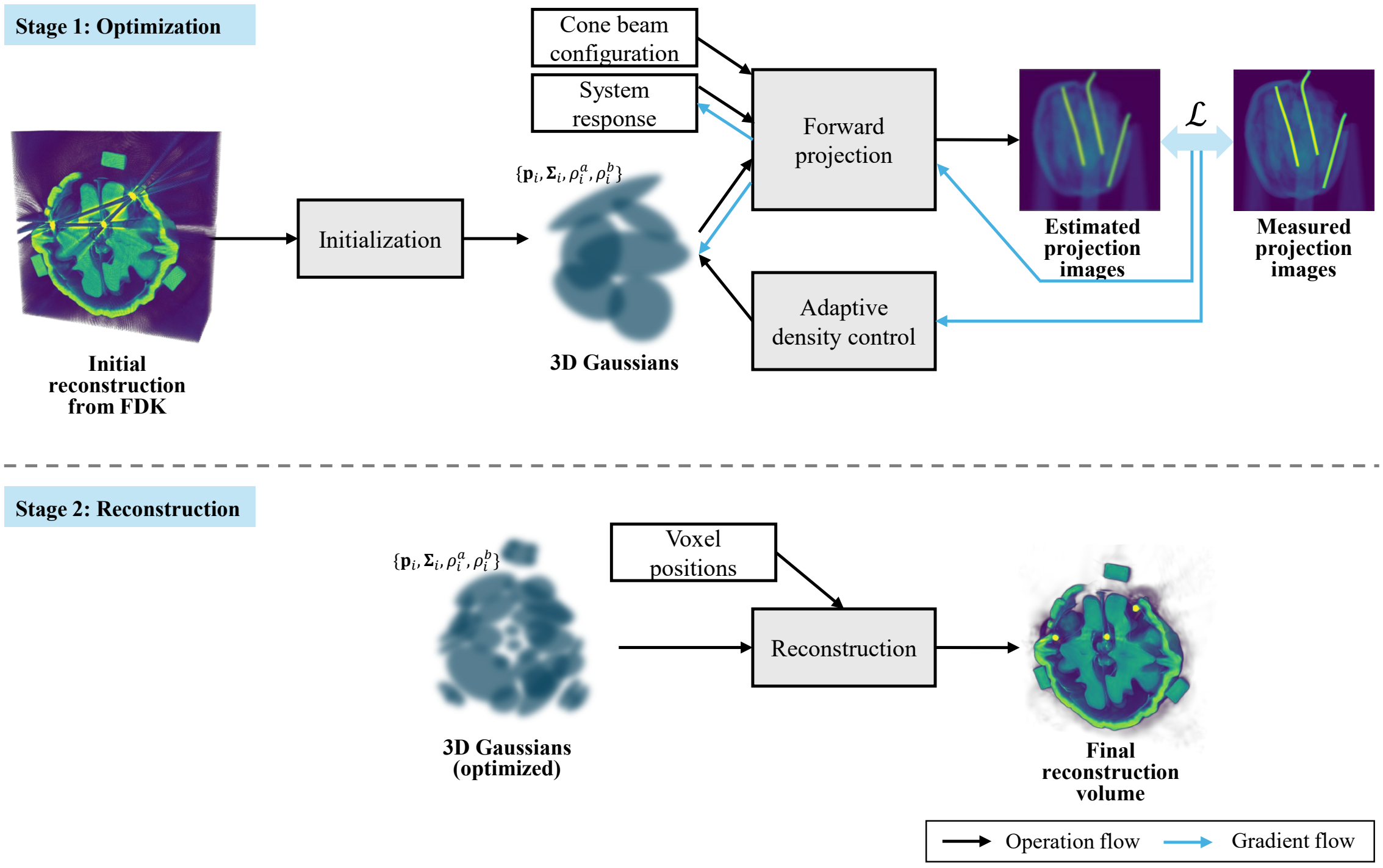


**Figure 1:** *Overview of our optimization pipeline.*

## 2. Additional Ablation Results

Figure 2 provides additional qualitative results that support the ablation experiments reported by Table 5 in the main paper. The numbers shown in the upper-right corners indicate the corresponding 3D PSNR values. As shown in this figure, the baseline reconstruction is noticeably blurred and suffers from severe metal artifacts. When only our response model is applied, finer details are recovered; however, prominent metal artifacts remain since the beam-hardening is not taken into account. Applying only the attenuation model significantly reduces these artifacts, although minor residual artifacts persist near metallic regions. When both models are jointly employed, metal artifacts are further suppressed and the overall intensity level closely matches the ground-truth volume, resulting in the highest PSNR among all ablation settings.

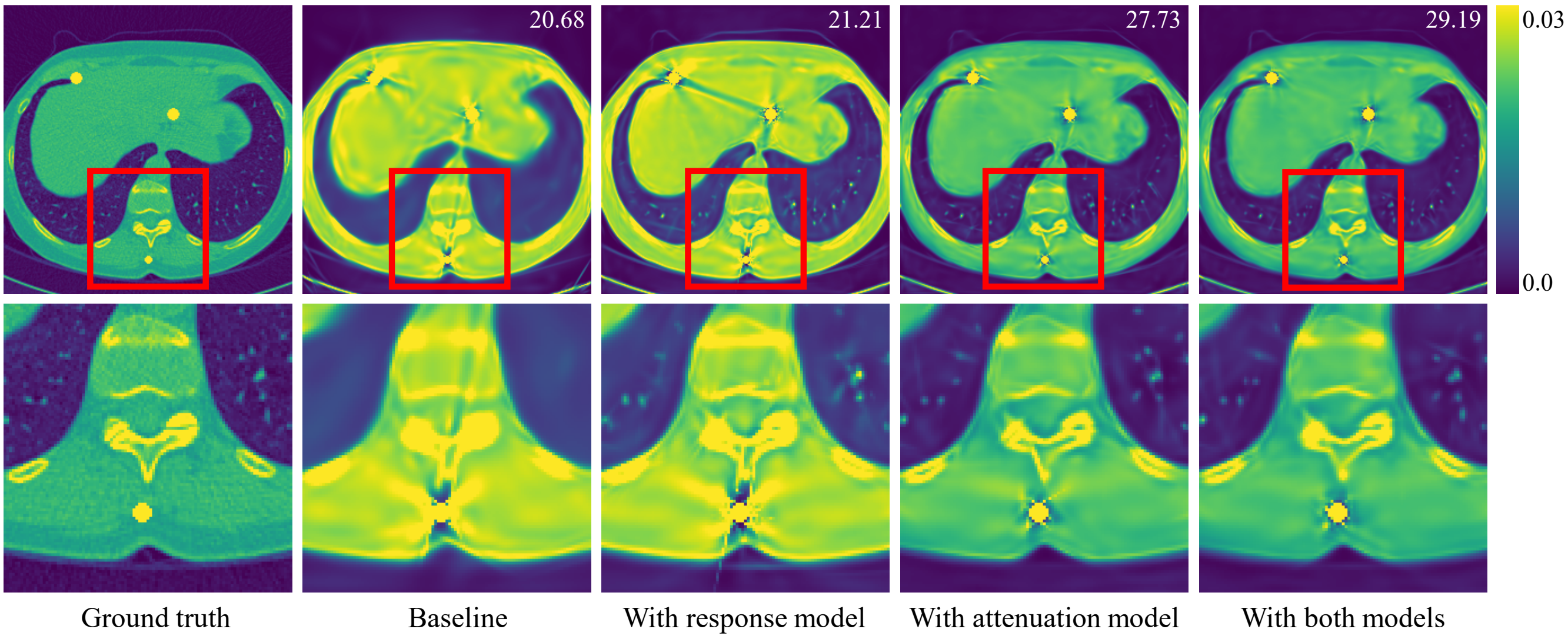


**Figure 2:** *Qualitative results of the ablations reported by Table 5 in the main paper.*

## 3. Derivation of Backward Propagation

We implement a custom CUDA pipeline to enable automatic differentiation of the parameters in the discrete polychromatic forward projection model using the Gaussian primitives as Equation (1):

$$P(\hat{\mathbf{x}}) = \sum_{i}^{M} s_i G_i(\hat{\mathbf{x}}) \rho_i^a - \log \sum_{k=0}^{N-1} \eta_k \exp\left( - \frac{1}{\left( \left( \frac{2k}{N-1} - 1 \right) \gamma + 1 \right)^3} \sum_{i}^{M} s_i G_i(\hat{\mathbf{x}}) \rho_i^b \right), \tag{1}$$

where $s_i = \sqrt{\frac{2\pi|\tilde{\Sigma}_i|}{|\Sigma_i|}}$, $G_i(\hat{\mathbf{x}}) = \exp\left(-\frac{1}{2}(\hat{\mathbf{x}} - \hat{\mathbf{p}}_i)^T \tilde{\Sigma}_i^{-1} (\hat{\mathbf{x}} - \hat{\mathbf{p}}_i)\right)$, $\sum_{k=0}^{N-1} \eta_k = 1$.

$R^2$-Gaussian [ZLC*24] consists of the `Preprocessor` that transforms the position and shape of each Gaussian, and the `Rasterizer` that computes the intensity projected onto each detector pixel. While the `Preprocessor` reuses the original implementation, we design a

new `Rasterizer` model and manually derive the gradients with respect to each intermediate parameter as Equations (2)–(6):

$$\left.\frac{d\mathcal{L}}{d\rho_i^a}\right|_{\hat{\mathbf{x}}} = \frac{d\mathcal{L}}{dP(\hat{\mathbf{x}})} \cdot \frac{dP(\hat{\mathbf{x}})}{d\rho_i^a} = \frac{d\mathcal{L}}{dP(\hat{\mathbf{x}})} \cdot \left(s_i G_i(\hat{\mathbf{x}})\right), \tag{2}$$

$$\left.\frac{d\mathcal{L}}{d\rho_i^b}\right|_{\hat{\mathbf{x}}} = \frac{d\mathcal{L}}{dP(\hat{\mathbf{x}})} \cdot \frac{dP(\hat{\mathbf{x}})}{d\rho_i^b} = \frac{d\mathcal{L}}{dP(\hat{\mathbf{x}})} \cdot \left(-\frac{1}{A(\hat{\mathbf{x}})} s_i G_i(\hat{\mathbf{x}}) \sum_{k=0}^{N-1} \left(B_k(\hat{\mathbf{x}}) C_k\right)\right), \tag{3}$$

$$\left.\frac{d\mathcal{L}}{ds_i}\right|_{\hat{\mathbf{x}}} = \frac{d\mathcal{L}}{dP(\hat{\mathbf{x}})} \cdot \frac{dP(\hat{\mathbf{x}})}{ds_i} = \frac{d\mathcal{L}}{dP(\hat{\mathbf{x}})} \cdot \left(\rho_i^a G_i(\hat{\mathbf{x}}) - \frac{1}{A(\hat{\mathbf{x}})} \rho_i^b G_i(\hat{\mathbf{x}}) \sum_{k=0}^{N-1} \left(B_k(\hat{\mathbf{x}}) C_k\right)\right), \tag{4}$$

$$\left.\frac{d\mathcal{L}}{dG_i}\right|_{\hat{\mathbf{x}}} = \frac{d\mathcal{L}}{dP(\hat{\mathbf{x}})} \cdot \frac{dP(\hat{\mathbf{x}})}{dG_i(\hat{\mathbf{x}})} = \frac{d\mathcal{L}}{dP(\hat{\mathbf{x}})} \cdot \left(\rho_i^a s_i - \frac{1}{A(\hat{\mathbf{x}})} \rho_i^b s_i \sum_{k=0}^{N-1} \left(B_k(\hat{\mathbf{x}}) C_k\right)\right), \tag{5}$$

$$\left.\frac{d\mathcal{L}}{d\eta_k}\right|_{\hat{\mathbf{x}}} = \frac{d\mathcal{L}}{dP(\hat{\mathbf{x}})} \cdot \frac{dP(\hat{\mathbf{x}})}{d\eta_k} = \frac{d\mathcal{L}}{dP(\hat{\mathbf{x}})} \cdot \left(-\frac{1}{A(\hat{\mathbf{x}})} \frac{B_k(\hat{\mathbf{x}})}{\eta_k}\right), \tag{6}$$

where $A(\hat{\mathbf{x}}) = \sum_{k=0}^{N-1} B_k(\hat{\mathbf{x}})$, $B_k(\hat{\mathbf{x}}) = \eta_k \exp\left(-C_k \sum_i^M \rho_i^b s_i G_i(\hat{\mathbf{x}})\right)$, $C_k = -\frac{1}{\left(\left(\frac{2k}{N-1}-1\right)\gamma+1\right)^3}$. All these equations are then implemented in the CUDA backward function. The final gradient with respect to $\eta_k$ is obtained by summing the pixel-wise partial derivatives:

$$\frac{d\mathcal{L}}{d\eta_k} = \sum_{\hat{\mathbf{x}}} \left.\frac{d\mathcal{L}}{d\eta_k}\right|_{\hat{\mathbf{x}}} = \sum_{\hat{\mathbf{x}}} \frac{d\mathcal{L}}{dP(\hat{\mathbf{x}})} \cdot \left(-\frac{1}{A(\hat{\mathbf{x}})} \cdot \frac{B_k(\hat{\mathbf{x}})}{\eta_k}\right) \tag{7}$$

## 4. Comparison Implementation Detail

Since both Polyner [WCW*23] and Park et al. [PSJ] are originally designed for the fan-beam geometry, we extend their implementations to support the cone-beam configuration for fair comparison with our method. Specifically, we constrain the sampling range that is originally set to (-SOD,SOD) where SOD is Source-to-Object Distance, to a narrower interval that better matches the actual object region to reduce sensitivity to variations across different datasets. To improve the reconstruction quality under the cone-beam setting, which is more challenging to reconstruct a volume due to its 3D geometry of beams, we also increase the depth of the network and expand the hash encoding parameters used in the reconstruction. As the official implementation of Park et al. is not publicly available, we instead adapt the extended cone-beam implementation of Polyner and fine-tune several network and training-related parameters for fair comparison. For learning-based methods such as ACDNet, DICDNet, and OSCNet, we employ publicly available pre-trained models. As these baselines only support metal artifact reduction on 2D slices, we first reconstruct a 3D volume using FDK and then apply each model independently to the individual slices.

## 5. Geometry Configuration

The geometry configurations for the datasets are provided in Tables 1–2. To validate the robustness of our method, we conduct experiments under various geometric setups.

## 6. GPU Memory Usage

We report the memory footprint of the baseline and our method in Table 3. Although our model is more complex than the baseline, its memory requirement increases only marginally, indicating that the additional polychromatic features do not compromise memory efficiency.

## 7. Additional Discussion

We compute the PSNR and SSIM for the 3D volume to compare the reconstruction performance on the synthetic dataset. For the PSNR calculation, the ground truth (GT) is determined based on the center energy, and following the evaluation methods of existing metal artifact reduction techniques [WCW*23] and [PSJ], PSNR is measured for all regions except for the metal mask area. Since there is no proper GT available for the real dataset, we do not conduct quantitative evaluations, but qualitative comparison evaluations are done instead.

## 8. Additional Results

Figure 3 presents the objects for which we provide additional results. Figures 4–6 show qualitative comparisons of our method against FDK [FDK84] and the state-of-the-art joint reconstruction and metal artifact reduction methods: Polyner [WCW*23], and Park et al. [PSJ].

**Table 1:** *The scanning configuration of our synthetic datasets.*

| **Parameters** | **Synthetic Dataset** | | |
|---|---|---|---|
| | Broccoli | Lung | Teeth |
| Source voltage (kV) | 90 | ← | ← |
| Aluminium filter (mm) | 0.5 | ← | ← |
| Metal material | Aluminum | Steel | Titanium |
| Volume resolution | 256×256×256 | ← | ← |
| Volume size (mm) | 50.0×50.0×50.0 | ← | ← |
| Angle range (°) | [0, 360) | ← | ← |
| Number of angles | 720 | ← | ← |
| Detector resolution | 512×512 | ← | ← |
| Detector size (mm) | 145.485 | ← | ← |
| Source-to-object distance (mm) | 200.962 | ← | ← |
| Source-to-detector distance (mm) | 501.309 | ← | ← |

**Table 2:** *The scanning configuration of our real datasets.*

| **Parameters** | **Real Dataset** | | | | |
|---|---|---|---|---|---|
| | Walnut | Metal Rods | Chicken (Wire) | Bell Pepper (Rivet) | Broccoli (Rivet) |
| Source voltage (kV) | 90 | ← | ← | ← | ← |
| Source current ($\mu$A) | 166 | ← | 300 | ← | ← |
| Aluminum filter (mm) | 1.0 | 0.5 | 1.0 | ← | ← |
| Metal material | Steel | Aluminum | Steel | ← | ← |
| Volume resolution | 256×256×256 | ← | ← | ← | ← |
| Volume size (mm) | 40.0×40.0×40.0 | 60.0×60.0×60.0 | 120.0×120.0×120.0 | ← | ← |
| Angle range (°) | [0, 360) | ← | ← | ← | ← |
| Number of angles | 720 | ← | ← | ← | ← |
| Detector resolution | 512×512 | ← | ← | ← | ← |
| Detector size (mm) | 104.773 | 145.485 | 229.902 | 145.485 | 229.902 |
| Source-to-object distance (mm) | 200.962 | ← | 334.942 | ← | ← |
| Source-to-detector distance (mm) | 501.309 | ← | ← | ← | ← |

**Table 3:** *GPU Memory usage (in MB) comparison between the baseline [ZLC*24] and our method across different scenes.*

| Type | Scene | Baseline (MB) | Ours (MB) |
|---|---|---|---|
| Synthetic | Broccoli | 1212 | 1402 |
| | Lung | 1242 | 1404 |
| | Teeth | 1272 | 1398 |
| Real | Walnut | 1308 | 1390 |
| | Metal Rods | 1330 | 1406 |
| | Chicken (Wire) | 1196 | 1382 |
| | Bell Pepper (Rivet) | 1210 | 1388 |
| | Broccoli (Rivet) | 1256 | 1340 |

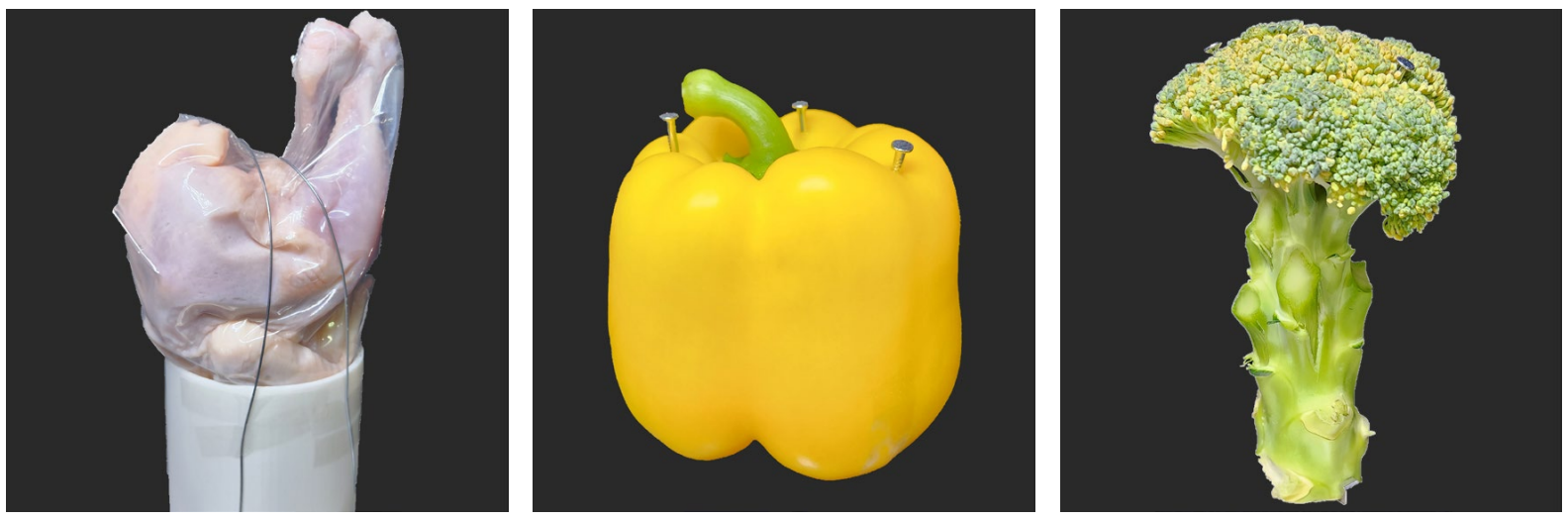

Chicken (Wire) Bell Pepper (Rivet) Broccoli (Rivet)

**Figure 3:** *The objects used for our additional results. The type of metallic objects used for each scene is shown in the parentheses.*

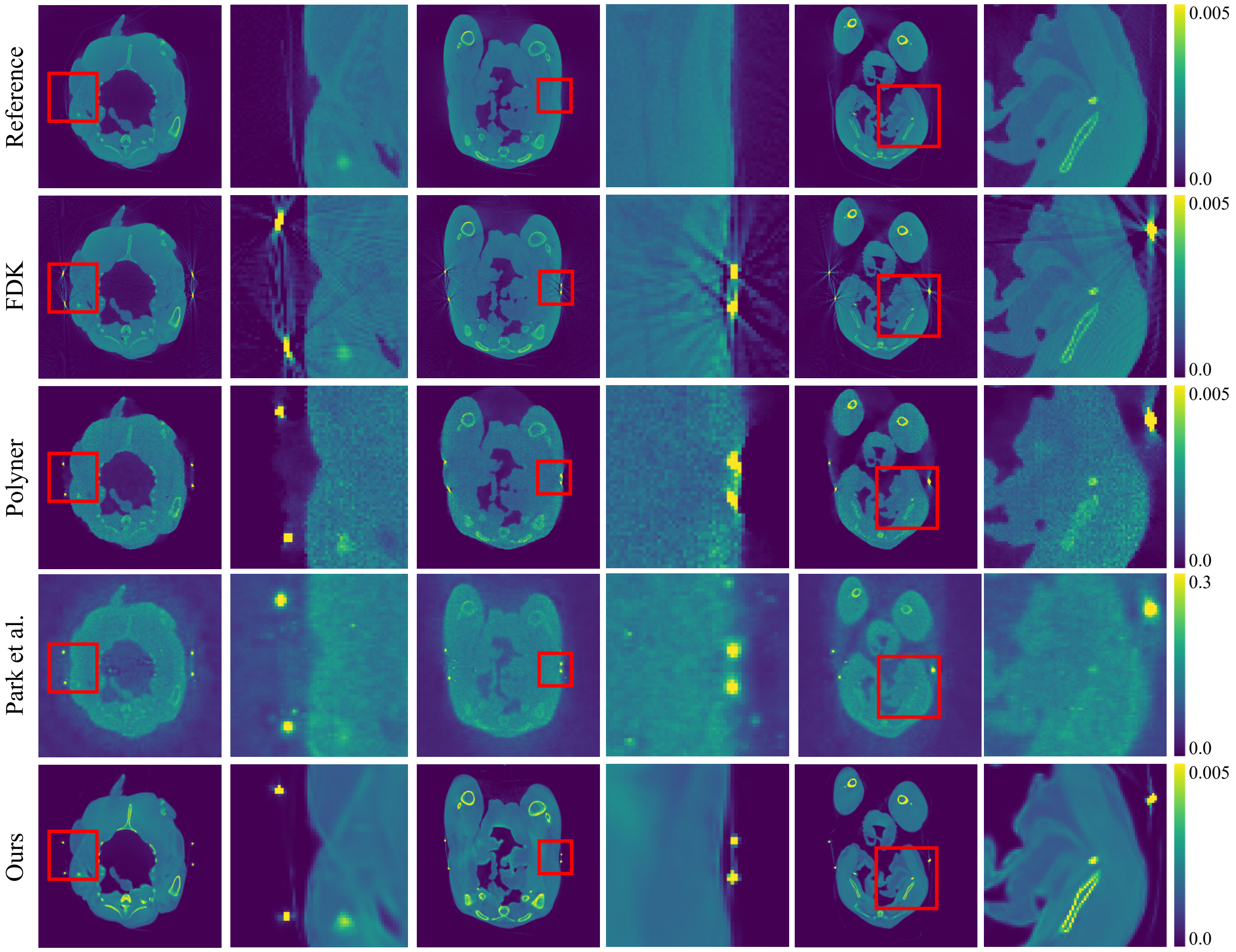


**Figure 4:** *Qualitative comparison on the real dataset: Chicken (Wire). The odd columns show the sampled slices from the reconstructed volumes by each algorithm, and the red boxes indicate the close-up regions. The even columns show the zoom-in patches from the corresponding slices.*

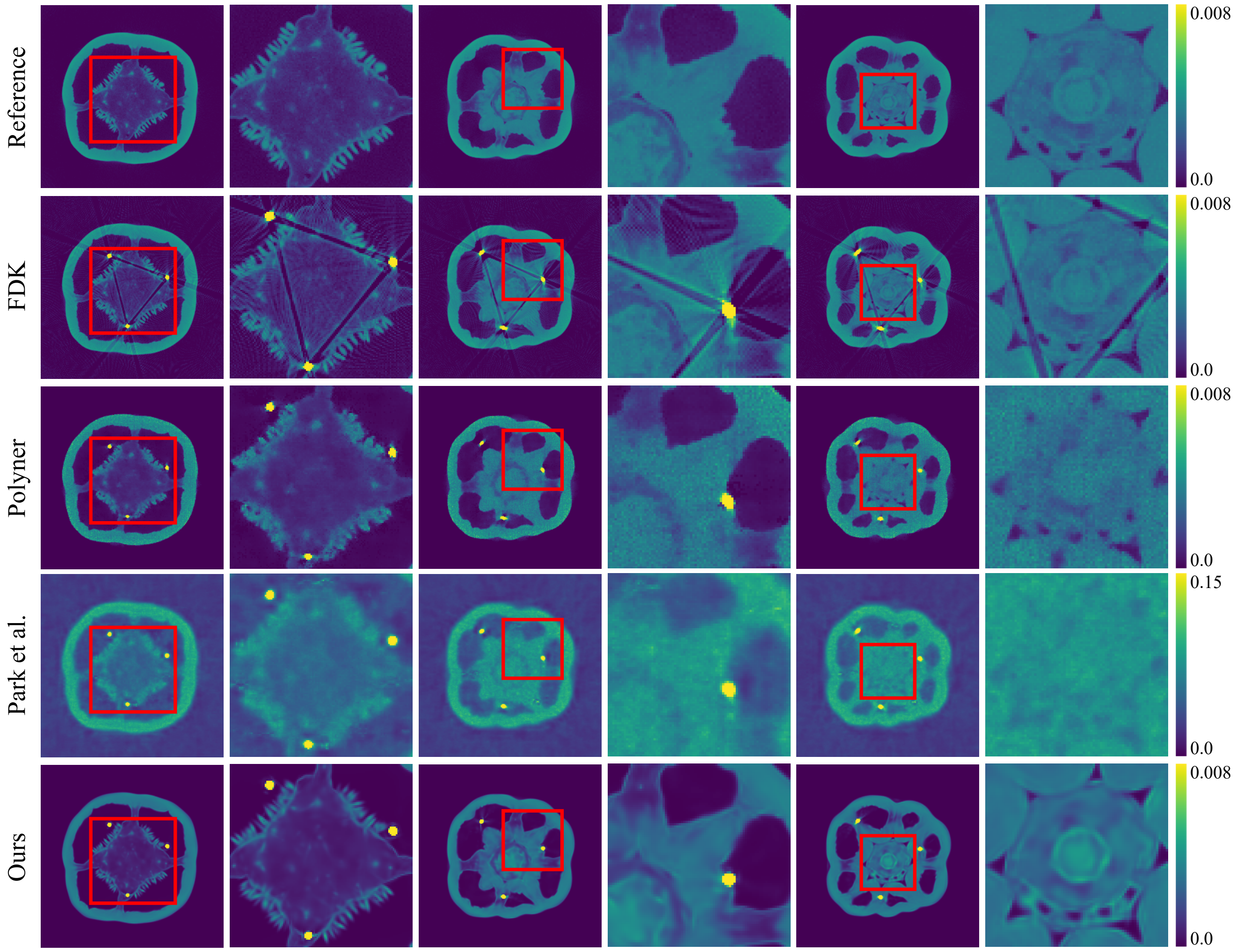


**Figure 5:** *Qualitative comparison on the real dataset: Bell Pepper (Rivet). The odd columns show the sampled slices from the reconstructed volumes by each algorithm, and the red boxes indicate the close-up regions. The even columns show the zoom-in patches from the corresponding slices.*

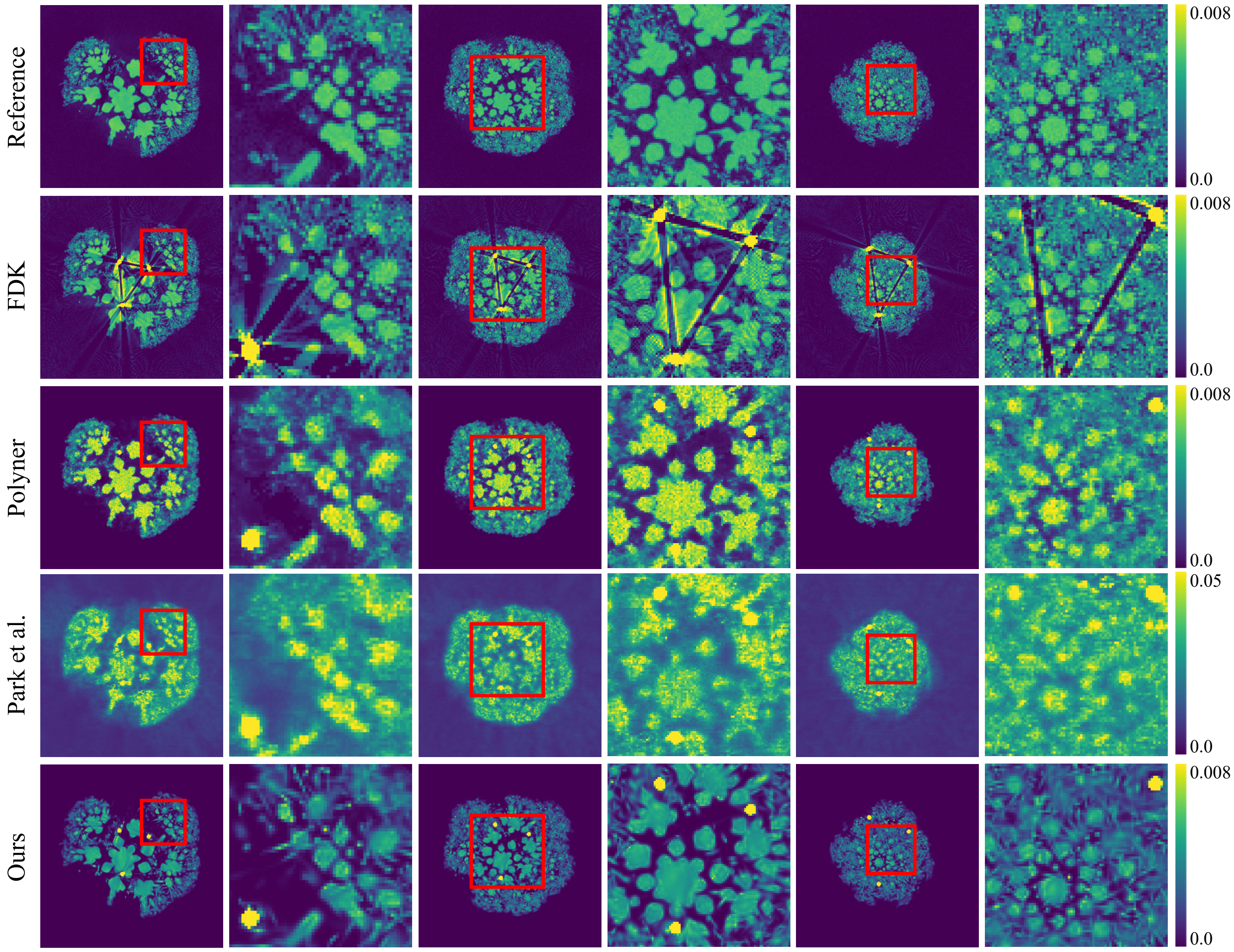


**Figure 6:** *Qualitative comparison on the real dataset: Broccoli (Rivet). The odd columns show the sampled slices from the reconstructed volumes by each algorithm, and the red boxes indicate the close-up regions. The even columns show the zoom-in patches from the corresponding slices.*


\clearpage 
\pagestyle{plain} 
	
\end{document}